# A Rigorously Bayesian Beam Model and an Adaptive Full Scan Model for Range Finders in Dynamic Environments

**Tinne De Laet** TINNE.DELAET@MECH.KULEUVEN.BE
**Joris De Schutter** JORIS.DESCHUTTER@MECH.KULEUVEN.BE
**Herman Bruyninckx** HERMAN.BRUYNINCKX@MECH.KULEUVEN.BE
*Department of Mechanical Engineering*
*Katholieke Universiteit Leuven*
*Celestijnenlaan 300B, box 2420, 3001 Heverlee, Belgium*

## Abstract

This paper proposes and experimentally validates a Bayesian network model of a range finder adapted to dynamic environments. All modeling assumptions are rigorously explained, and all model parameters have a physical interpretation. This approach results in a transparent and intuitive model. With respect to the state of the art beam model this paper: (i) proposes a different functional form for the probability of range measurements caused by unmodeled objects, (ii) intuitively explains the discontinuity encountered in the state of the art beam model, and (iii) reduces the number of model parameters, while maintaining the same representational power for experimental data. The proposed beam model is called RBBM, short for Rigorously Bayesian Beam Model. A maximum likelihood and a variational Bayesian estimator (both based on expectation-maximization) are proposed to learn the model parameters.

Furthermore, the RBBM is extended to a full scan model in two steps: first, to a full scan model for static environments and next, to a full scan model for general, dynamic environments. The full scan model accounts for the dependency between beams and adapts to the local sample density when using a particle filter. In contrast to Gaussian-based state of the art models, the proposed full scan model uses a sample-based approximation. This sample-based approximation enables handling dynamic environments and capturing multi-modality, which occurs even in simple static environments.

## 1. Introduction

In a probabilistic approach, inaccuracies are embedded in the stochastic nature of the model, particularly in the conditional probability density representing the measurement process. It is of vital importance that all types of inaccuracies affecting the measurements are incorporated in the probabilistic sensor model. Inaccuracies arise from sensor limitations, noise, and the fact that most complex environments can only be represented and perceived in a limited way. The dynamic nature of the environment in particular is an important source of inaccuracies. This dynamic nature results from the presence of unmodeled and possibly moving objects and people.

This paper proposes a probabilistic range finder sensor model for dynamic environments. Range finders, which are widely used in mobile robotics, measure the distances $\boldsymbol{z}$ to objects in the environment along certain directions $\boldsymbol{\theta}$ relative to the sensor. We derive the sensor model in a form suitable for mobile robot localization, i.e.: $P(Z = z \mid X = x, M = m)^1$, where $Z$ indicates the measured range, $X$ the position of the mobile robot (and of the sensor mounted on it), and $M$ the environment map. The presented model is however useful in other applications of range sensors as well.

First, this paper derives a probabilistic sensor model for one beam of a range finder, i.e. the *beam model*. In particular, this paper gives a rigorously Bayesian derivation using a Bayesian network model while stating all model assumptions and giving a physical interpretation for all model parameters. The obtained model is named RBBM, short for Rigorously Bayesian Beam Model. The innovations of the presented approach are (i) to introduce *extra state variables* $A = a$ for the positions of unmodeled objects in the probabilistic sensor model $P(z \mid x, m, a)$, and (ii) *to marginalize out* these extra state variables from the total probability *before* estimation. The latter is required because extra variables (exponentially!) increase the computational complexity of state estimation while in a lot of applications estimating the position of unmodeled objects is not of primary interest. In summary, the marginalization avoids the increase in complexity to infer the probability distributions $P(x)$ and $P(m)$, while maintaining the modeling of the dynamic nature of the environment.

This paper furthermore presents a maximum-likelihood and a variational Bayesian estimator (both based on expectation-maximization) to learn the model parameters of the RBBM.

Next, the paper presents an extension of the RBBM to a full scan model i.e.: $P(\boldsymbol{z} \mid \boldsymbol{\theta}, x, m)$ where $\boldsymbol{z}$ and $\boldsymbol{\theta}$ contain all the measured distances and beam angles, respectively. This full scan model accounts for the dependency between beams and adapts to the local sample density when using a particle filter. In contrast to Gaussian-based state of the art models, the proposed full scan model uses a sample-based approximation. The sample-based approximation allows us to capture the multi-modality of the full scan model, which is shown to occur even in simple static environments.

### 1.1 Paper Overview

This paper is organized as follows. Section 2 gives an overview of the related work. Section 3 (i) presents a Bayesian beam model for range finders founded on Bayesian networks, the RBBM, (ii) mathematically derives an analytical formula for the probabilistic sensor model while clearly stating all assumptions, (iii) provides useful insights in the obtained beam model and (iv) shows that the obtained analytical sensor model agrees with the proposed Bayesian network. Section 4 presents a maximum likelihood and a variational Bayesian estimator (both based on expectation-maximization) to learn the model parameters. In Section 5 the model parameters of the RBBM are learned from experimental data and the resulting model is compared with the state of the art beam model proposed by Thrun, Burgard, and Fox (2005), further on called Thrun's model. Section 6 extends the RBBM to an adaptive full scan model for dynamic environments. Section 7 discusses the obtained RBBM and adaptive full scan model and compares them with previously proposed range finder sensor models.

1. To simplify notation, the explicit mention of the random variable in the probabilities is omitted whenever possible, and replaced by the common abbreviation $P(x)$ instead of writing $P(X = x)$.

## 2. Related Work

Three basic approaches to deal with dynamic environments exist in the literature (Fox, Burgard, & Thrun, 1999; Thrun et al., 2005): state augmentation, adapting the sensor model and outlier detection.

In *state augmentation* the latent states, e.g. the position of moving objects and people in the environment, are included in the estimated states. Wang, Thorpe, and Thrun (2003) developed an algorithm 'SLAM with DATMO', short for SLAM with the detection and tracking of moving objects. State augmentation however is often infeasible since the computational complexity of state estimation increases exponentially with the number of independent state variables to estimate. A closely related solution consists of adapting the map according to the changes in the environment. Since such approaches assume that the environment is almost static, they are unable to cope with real dynamics as in populated environments (Fox et al., 1999). A more recent, related approach proposed by Wolf and Sukhatme (2004) maintains two coupled occupancy grids of the environment, one for the static map and one for the moving objects, to account for environment dynamics.

Probabilistic approaches are to some extent robust to unmodeled dynamics, since they are able to deal with sensor noise. In such approaches however, the sensor noise should reflect the real uncertainty due to the unmodeled dynamics of the environment. Therefore, a second approach for dealing with dynamic environments is to *adapt the sensor model* to correctly reflect situations in which measurements are affected by the unmodeled environment dynamics. Fox et al. (1999) show that such approaches are only capable to model such noise on average, and, while these approaches work reliably with occasional sensor blockage, they are inadequate in situations where more than fifty percent of the measurements are corrupted.

To handle measurement corruption more effectively, an approach based on *outlier detection* can be used. This approach uses an adapted sensor model, as explained in the previous paragraph. The idea is to investigate the cause of a sensor measurement and to reject measurements that are likely to be affected by unmodeled environment dynamics. Hähnel, Schulz, and Burgard (2003a) and Hähnel, Triebel, Burgard, and Thrun (2003b) studied the problem of performing SLAM in environments with many moving objects using the EM algorithm for filtering out affected measurements. By doing so, they were able to acquire maps in the environment where conventional SLAM techniques failed. Fox et al. (1999) propose two different kinds of filters: an entropy filter, suited for an arbitrary sensor, and a distance filter, designed for proximity sensors. These filters detect whether a measurement is corrupted or not, and discard sensor readings resulting from objects that are not contained in the map.

This paper focuses on (sonar and laser) range finders, whose physical principle is the emission of a sound or light wave, followed by the recording of its echo. Highly accurate sensor models would include *physical* parameters such as surface curvature and material absorption coefficient. These parameters are however difficult to estimate robustly in unstructured environments. Hence, the literature typically relies on purely basic *geometric* models.

The range finder sensor models available from the literature are traditionally divided in three main groups: feature-based approaches, beam-based models and correlation-based methods. *Feature-based approaches* extract a set of features from the range scan and match them to features contained in an environmental model. *Beam-based models*, also known as ray cast models, consider each distance measurement along a beam as a separate range measurement. These models represent the one-dimensional distribution of the distance measurement by a parametric function, which depends on the expected range measurement in the respective beam directions. In addition, these models are closely linked to the geometry and the physics involved in the measurement process. They often result in overly peaked likelihood functions due to the underlying assumption of independent beams. The last group of range finder sensor models, *correlation-based methods*, build local maps from consecutive scans and correlate them with a global map. The simple and efficient likelihood field models or end point model (Thrun, 2001) are related to these correlation-based methods. Plagemann, Kersting, Pfaff, and Burgard (2007) nicely summarize the advantages and drawbacks of the different range finder sensor models.

Range finder sensor models can also be classified according to whether they use discrete geometric grids (Hähnel et al., 2003a, 2003b; Fox et al., 1999; Burgard, Fox, Hennig, & Schmidt, 1996; Moravec, 1988) or continuous geometric models (Thrun et al., 2005; Choset, Lynch, Hutchinson, Kantor, Burgard, Kavraki, & Thrun, 2005; Pfaff, Burgard, & Fox, 2006). Moravec proposed non-Gaussian measurement densities over a discrete grid of possible distances measured by sonar; the likelihood of the measurements has to be computed for all possible positions of the mobile robot at a given time. Even simplified models (Burgard et al., 1996) in this approach turned out to be computationally too expensive for real-time application. Therefore, Fox et al. proposed a beam model consisting of a mixture of *two* physical causes for a measurement: a hit with an object in the map, or with an object not yet modeled in the map. The last cause accounts for the dynamic nature of the environment. An analogous mixture (Thrun et al., 2005; Choset et al., 2005) adds two more physical causes: a sensor failure and an unknown cause resulting in a 'max-range' measurement and a 'random' measurement, respectively. While Thrun et al. and Pfaff et al. use a continuous model, Choset et al. present the discrete analog of the mixture, taking into account the limited resolution of the range sensor. Pfaff et al. extend the basic mixture model for use in Monte Carlo localization. To overcome problems due to the combination of the limited representational power and the peaked likelihood of the accurate range finder, they propose an *adaptive* likelihood model. The likelihood model is smooth during global localization and more peaked during tracking.

Recently, different researchers tried to tackle the problems associated with beam-based models, caused by the independence assumptions between beams. Plagemann et al. (2007) propose a sensor model for the full scan. The model treats the sensor modeling task as a non-parametric Bayesian regression problem, and solves it using Gaussian processes. It is claimed that the Gaussian beam processes combine the advantages of the beam-based and the correlation-based models. Due to the underlying assumption that the measurements are jointly Gaussian distributed, the Gaussian beam processes are not suited to take into account the non-Gaussian uncertainty due to the dynamic nature of the environment. An alternative approach to handle the overly-peaked likelihood functions resulting from the traditional beam models is proposed by Pfaff, Plagemann, and Burgard (2007). A location-dependent full scan model takes into account the approximation error of the sample-based representation, and explicitly models the correlations between individual beams introduced by the pose uncertainty. The measurements are assumed to be jointly Gaussian distributed just as Plagemann et al. proposed. While Plagemann et al. represent the covariance matrix as a parametrized covariance function using Gaussian processes whose parameters are learned from data, Pfaff et al. learn the full covariance matrix being less restrictive in this manner. Despite the modeled correlation between beams, the measurements are still assumed to be jointly Gaussian distributed, which again limits the applicability in dynamic environments.

This paper proposes a rigorously Bayesian modeling of the probabilistic range sensor beam model for dynamic environments, referred to as RBBM. Similar to the work of Thrun et al. (2005) and Pfaff et al. (2006) the sensor model is derived for a continuous geometry. Unlike previous models of Thrun et al. (2005), Pfaff et al. (2006), Fox et al. (1999) and Choset et al. (2005), the mixture components are founded on a Bayesian modeling. This modeling makes use of probabilistic graphical models, in this case Bayesian networks. Such graphical models provide a simple way to visualize the structure of a probabilistic model, and can be used to design and motivate new models (Bishop, 2006). By inspection of the graph, insights of the model, including conditional independence properties are obtained. Next, inspired by the adaptive full scan models in the literature (Pfaff et al., 2006, 2007; Plagemann et al., 2007), the RBBM is extended to an adaptive full scan model. The underlying sample-based approximation of the full scan model, in contrast to the Gaussian-based approximation proposed by Pfaff et al. (2007) and Plagemann et al., enables handling dynamic environments and capturing multi-modality, which occurs even in simple static environments.

## 3. Beam Model

We model the probabilistic beam model $P\left(Z=z \mid X=x, M=m\right)$ for dynamic environments as a Bayesian network. We introduce extra state variables $A=a$ for the positions of unmodeled objects in the probabilistic sensor model $P\left(z \mid x, m, a\right)$. To prevent an exponential increase of the computational complexity of the state estimation due to the extra variables, these variables are marginalized out from the total probability *before* estimation. The marginalization:

$$P\left(z \mid x, m\right)=\int_{a} P\left(z \mid x, m, a\right) P\left(a\right) da, \tag{1}$$

avoids increasing the complexity to infer the conditional probability distributions of interest, $P\left(x\right)$ and $P\left(m\right)$, while it maintains the modeling of the dynamic nature of the environment. Section 3.1 explains which extra state variables are physically relevant, while Section 3.3 explains the marginalization of these extra state variables. Section 3.5 summarizes all assumptions and approximations. Finally, Section 3.6 provides useful insights in the obtained beam model, called RBBM, and in its derivation. Section 3.7 shows that the obtained analytical expression for the RBBM agrees with the proposed Bayesian network by means of a Monte Carlo simulation.

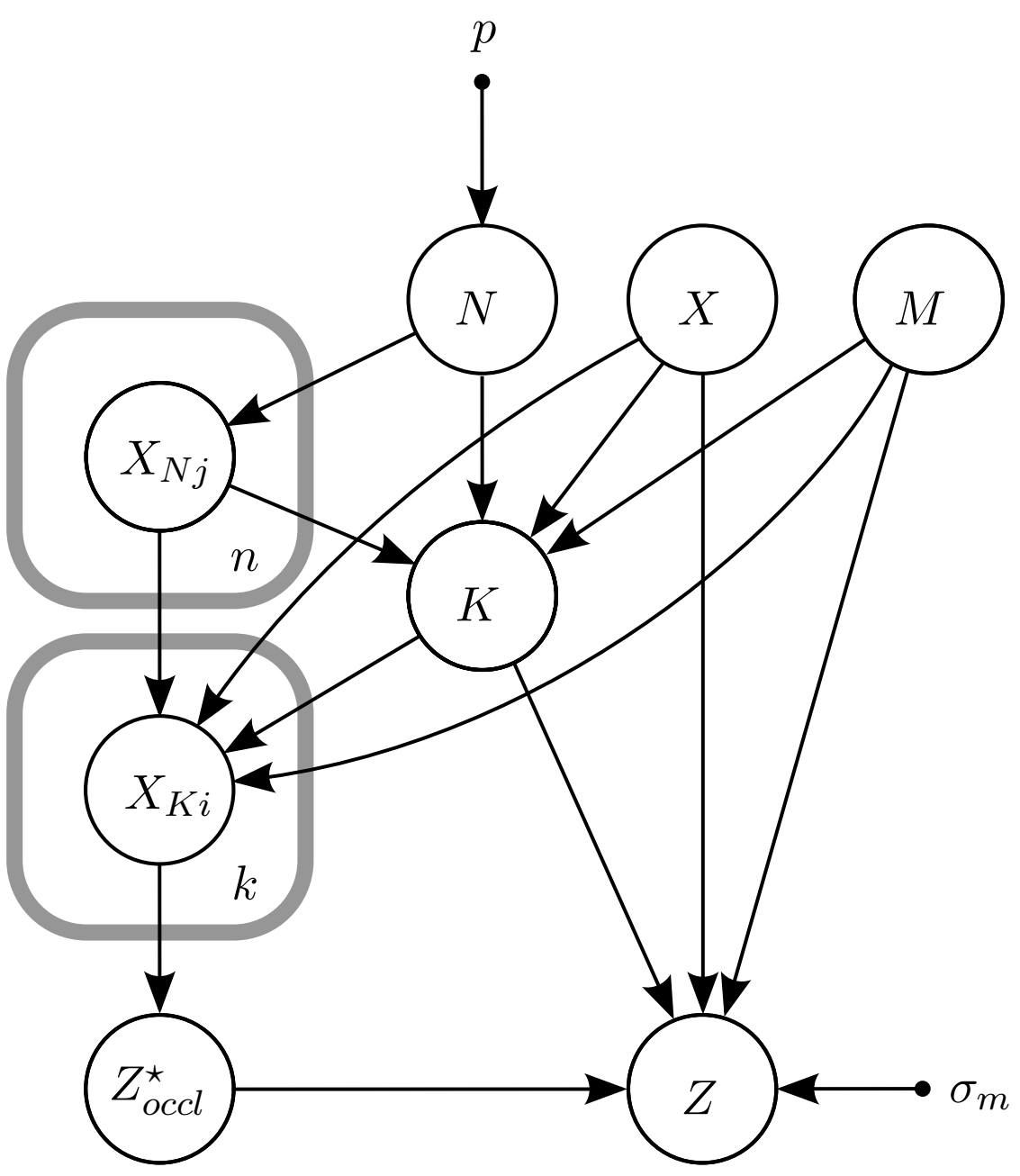

Figure 1: The Bayesian network for the probabilistic measurement model supplemented with the deterministic parameters represented by the smaller solid nodes. A compact representation with *plates* (the rounded rectangular boxes) is used. A plate represents a number, indicated in the lower right corner, of independent nodes of which only a single example is shown explicitly.

### 3.1 Bayesian Model

Bayesian networks graphically represent probabilistic relationships between variables in a mathematical model, to structure and facilitate probabilistic inference computations with those variables (Jensen & Nielsen, 2007; Neapolitan, 2004). A Bayesian network is defined as follows: (i) a set of nodes, each with an associated *random variable*, connected by *directed edges* forming a *directed acyclic graph (DAG)*; (ii) each discrete (continuous) random variable has a finite (infinite) set of mutually exclusive states; (iii) each random variable $A$ with parents $B_1, \ldots, B_N$ has a *conditional probability distribution* $P(A \mid B_1, \ldots, B_n)$ (known as conditional probability table in the case of discrete variables). Although the definition of Bayesian networks does not refer to causality, and there is no requirement that the directed edges represent causal impact, a well-known way of structuring variables for reasoning under uncertainty is to construct a graph representing causal relations (Jensen & Nielsen, 2007). In this case the graphical models are also known as *generative models* (Bishop, 2006), since they capture the causal process generating the random variables.

In this application, the range sensor ideally measures $z^\star$, the distance to the closest object in the map. An unknown number $n$ of unmodeled objects, possibly preventing the measurement of the closest object in the map, are however present in the environment. Depending on the position of the $j$th unmodeled object along the measurement beam, $x_{Nj}$, the unmodeled object occludes the map or not. The unmodeled object only occludes the map if it is located in front of the closest object contained in the map. $k$ is the total number of occluding objects out of the $n$ unmodeled objects. The positions of these occluding objects on the measurement beam are denoted by $\{x_{Ki}\}_{i=1:k}$. If the map is occluded by an unmodeled object, the range sensor will ideally measure $z^\star_{\text{occl}} = x_{Kc}$, with $x_{Kc}$ the position of the closest occluding object.

The following extra state variables, $a$ in Eq. (1), are included in the Bayesian model: $N$ is the discrete random variable indicating the unknown number of unmodeled objects in the environment; $X_{Nj}$ is the continuous random variable for the position of the $j$th unmodeled object on the measurement beam; $K$ is the discrete random variable indicating the number of objects occluding the measurement of the map; $X_{Ki}$ is the continuous random variable for the position of the $i$th occluding object on the measurement beam; and $Z^\star_{\text{occl}}$ is the continuous random variable indicating the ideal range measurement of the closest occluding object. Fig. 1 shows the Bayesian network for the probabilistic range finder sensor model with the variables $Z, X$ and $M$ that occur in the probabilistic sensor model (defined in Section 1), all the extra variables $N, X_N = \{X_{Nj}\}_{j=1:n}, K, X_K = \{X_{Ki}\}_{i=1:k}, Z^\star_{\text{occl}}$ and the model parameters $p$ and $\sigma_m$ (defined in Section 3.2).

The directed edges in the graphical model represent *causal relationships* between the variables. For example, $X$ and $M$ unambiguously determine the measured range $Z$ for a perfect sensor in the absence of unmodeled occluding objects. The number of occluding objects $K$ depends on the total number $N$ of unmodeled objects and their positions $X_N$ with respect to the measurement beam. $X$ and $M$ also have a causal impact on $K$: the larger the expected measurement $z^\star$, the higher the possibility that one or more unmodeled objects are occluding the modeled object corresponding to the expected measurement. The positions along the measurement beam $X_K$ of the occluding objects are equal to the positions of the $K$ of $N$ unmodeled objects occluding the map. Therefore, random variables $X_K$ are not only influenced by $K$ but also by $X_N$. Since the $K$ objects are occluding the map, their positions along the measurement beam are limited to the interval $[0, z^\star]$, so $X_K$ has a causal dependency on $X$ and $M$. The ideal measurement $z^\star_{\text{occl}}$ of an occluding object is the position of the occluding object closest to the sensor, so $Z^\star_{\text{occl}}$ depends on the positions $X_K$ of the occluding objects. Finally, measurement $Z$ also depends on the ideal measurement of the occluding object $Z^\star_{\text{occl}}$ and the number of occluding objects $K$. In case of occlusion ($k \geq 1$), $z^\star_{\text{occl}}$ is ideally measured, else (no occlusion, $k = 0$) $z^\star$ is ideally measured.

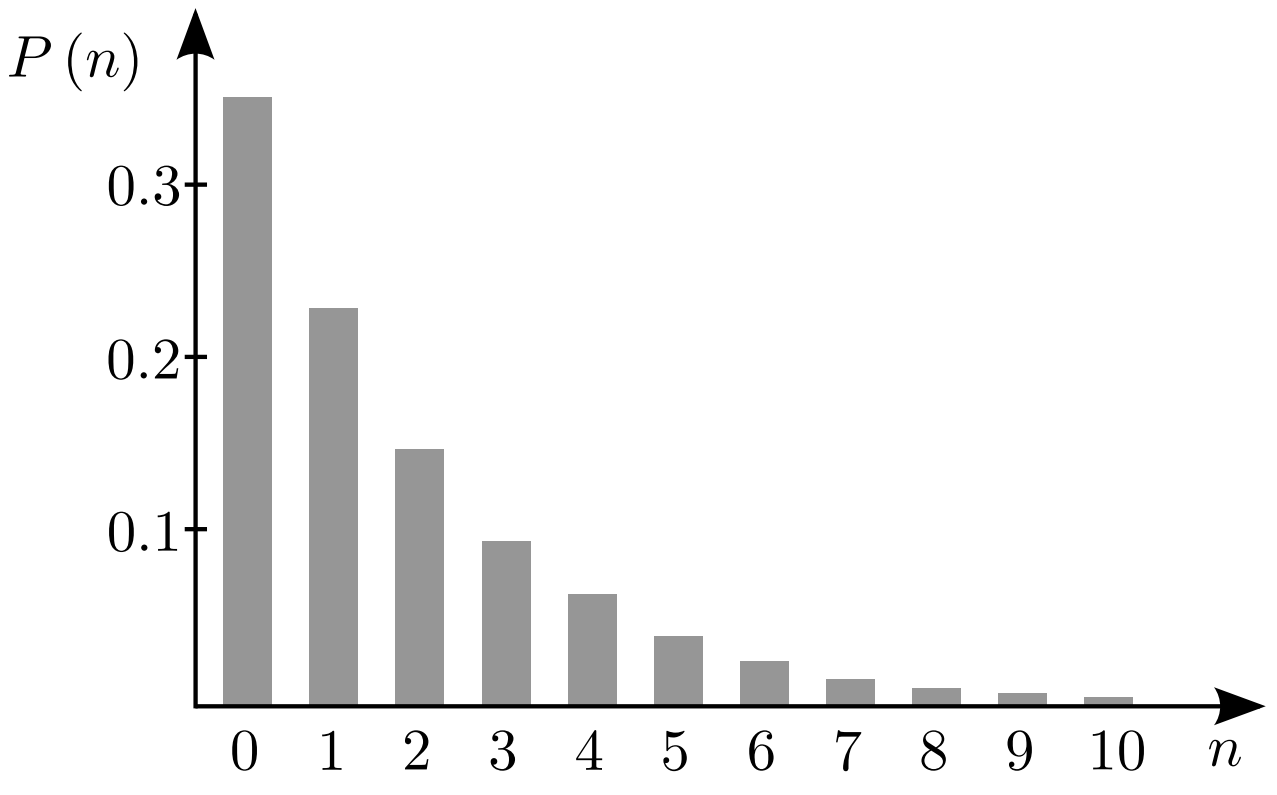

Figure 2: $P(n)$ (Eq. (2)) for $p = 0.65$.

### 3.2 Conditional Probability Distributions

Inferring the probability distribution of the extra state variables such as $P(n)$ is often infeasible. Marginalization of the extra state variables $Z, X, M, N, X_N, K, X_K, Z^\star_{\text{occl}}$ avoids the increase in complexity of the estimation problem, but still takes into account the dynamic nature of the environment. Marginalization requires the modelling of *all conditional probability tables and conditional probability distributions (pdf)* of each random variable conditionally on its parents.

First of all, some assumptions have to be made for $P(n)$. *Assume that the probability of the number of unmodeled objects decreases exponentially*, i.e. $P(n)$ is given by:

$$P(n) = (1 - p)\, p^n, \tag{2}$$

with $p$ a measure for the degree of appearance of unmodeled objects. More precisely, $p$ is the probability that at least one unmodeled object is present. While $p$ is indicated in Fig. 1, Fig. 2 shows the resulting distribution $P(n)$.

Secondly, *assume that nothing is known a priori about the position of the unmodeled objects along the measurement beam.* Hence each unmodeled object's position is assumed to be uniformly distributed over the measurement beam (Fig. 3):

$$P(x_{Nj}) = \begin{cases} \frac{1}{z_{\max}} & \text{if } x_{Nj} \leq z_{\max} \\ 0 & \text{otherwise,} \end{cases} \tag{3}$$

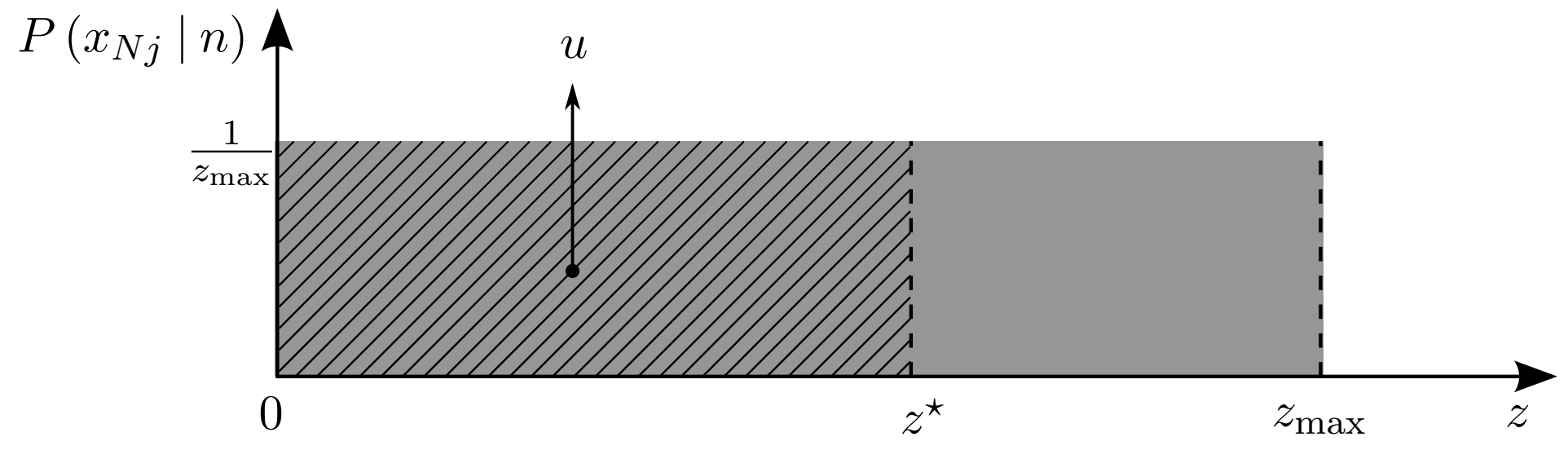

Figure 3: $P(x_{Nj} \mid n)$ (Eq. (3)).

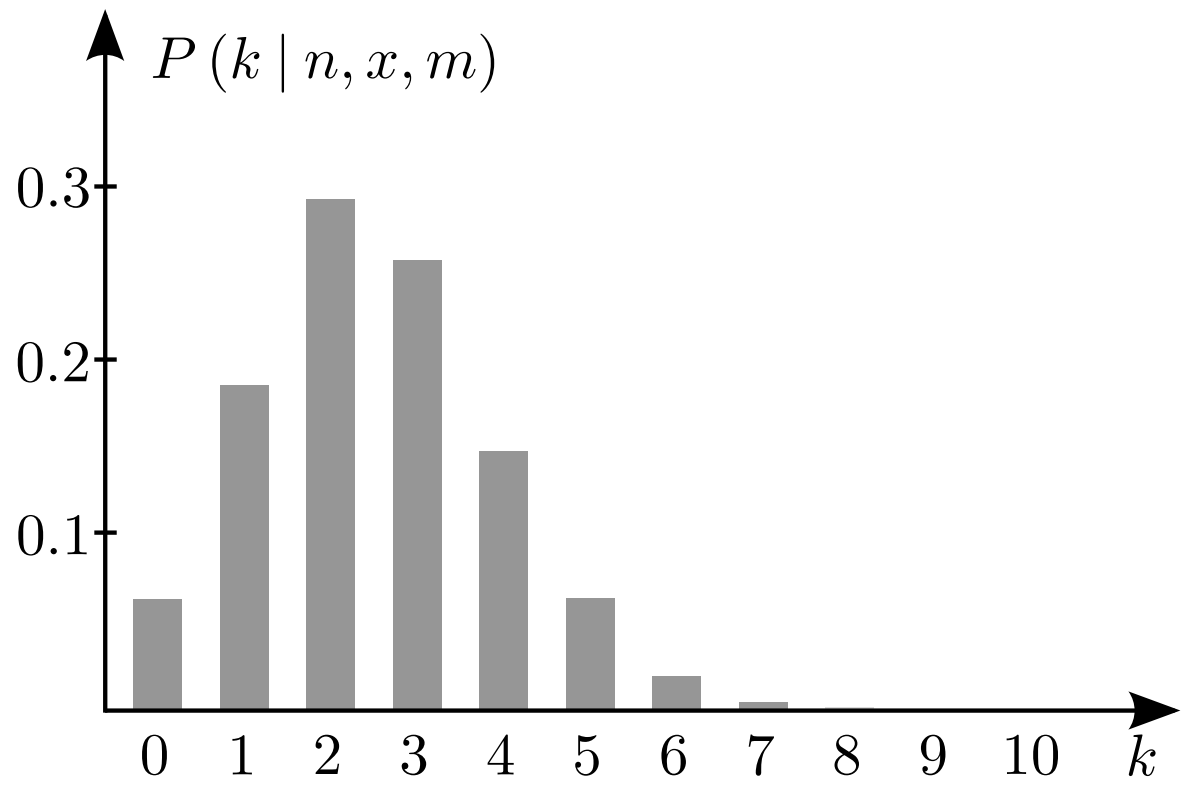

Figure 4: $P(k \mid n, x, m)$ (Eq. (5)) for $n = 10$ and $u = 0.25$.

with $z_{\max}$ the maximum range of the range sensor.

Thirdly, *assume the positions of the unmodeled objects are independent*:

$$P(x_N \mid n) = \prod_{j=1}^{n} P(x_{Nj}). \tag{4}$$

Next, an expression is needed for the conditional probability: $P(k \mid n, x_N, x, m)$, i.e. the probability that $k$ of the $n$ unmodeled objects are occluding the map $m$. An unmodeled object is occluding the map $m$ if it is located along the measurement beam and in front of the closest object in the map. It is straightforward to show that $P(k \mid n, x_N, x, m)$ is a binomial distribution:

$$P(k \mid n, x_N, x, m) = \begin{cases} \begin{pmatrix} n \\ k \end{pmatrix} u^k (1-u)^{n-k} & \text{if } k \leq n \\ 0 & \text{otherwise,} \end{cases} \tag{5}$$

where $u$ is the probability that an unmodeled object is occluding the map and $\begin{pmatrix} n \\ k \end{pmatrix} = \frac{n!}{(n-k)!k!}$ is the number of ways of selecting $k$ objects out of a total of $n$ objects. Fig. 4 shows this binomial distribution. Since it was assumed that the positions of the unmodeled objects were uniformly distributed, $u$, the probability that an unmodeled object is occluding the map is:

$$u = P\left(x_{Nj} < z^\star\right) = \frac{z^\star}{z_{\max}}, \tag{6}$$

as depicted in Fig. 3.

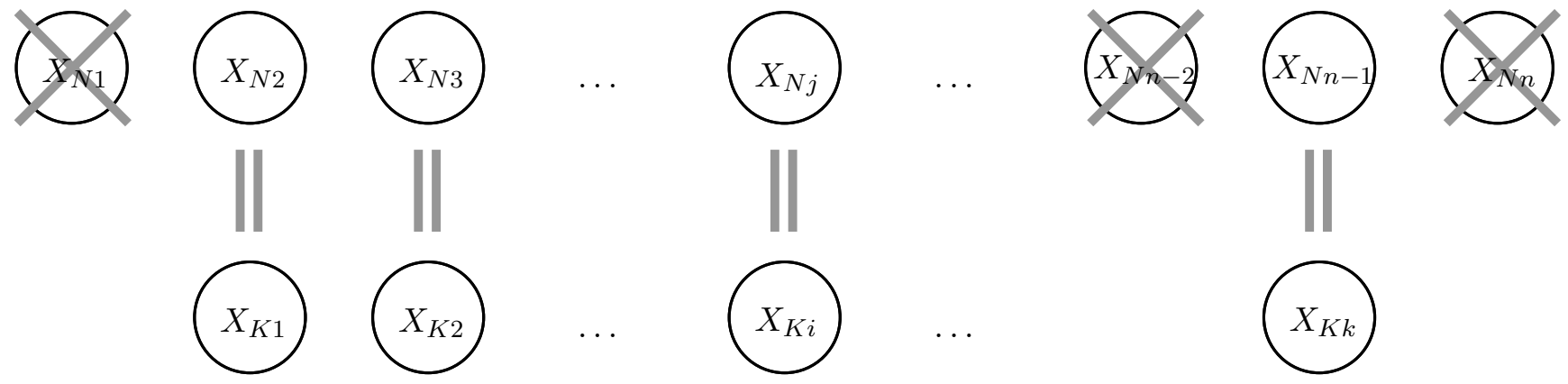

Figure 5: The selection scheme, where each cross eliminates an unmodeled object that is not occluding the map.

Furthermore, an analytical expression for $P\left(x_K \mid x_N, k\right)$ is necessary. The positions of the occluding objects $x_K$ are equal to the positions of the unmodeled objects $x_N$ that are occluding the map, as shown in Fig. 5. In other words, $x_{Ki}$ equals $x_{Nj}$ if and only if the unmodeled object is occluding the map, i.e. if $x_{Nj} \leq z^\star$:

$$P\left(x_{Ki} \mid x_{Nj}, k, x, m\right) = \begin{cases} \frac{1}{P\left(x_{Nj} \leq z^\star\right)} \delta\left(x_{Ki} - x_{Nj}\right) = \frac{z_{\max}}{z^\star} \delta\left(x_{Ki} - x_{Nj}\right) & \text{if } x_{Nj} \leq z^\star \\ 0 & \text{otherwise,} \end{cases} \tag{7}$$

with $\delta$ the Dirac function and $x_{Ki}$ the occluding object corresponding to $x_{Nj}$.

In case of occlusion, the range sensor ideally measures the distance to the closest occluding object $x_{Kc}$:

$$P\left(z^\star_{\text{occl}} \mid x_K\right) = \delta\left(z^\star_{\text{occl}} - x_{Kc}\right). \tag{8}$$

While range finders are truly quite deterministic since the measurements are to a great extent explainable by underlying physical phenomena such as specular reflections, inference, ... these underlying phenomena are complex and therefore costly to model. On top of these underlying phenomena additional uncertainty on the measurement is due to (i) uncertainty in the sensor position, (ii) inaccuracies of the world model and (iii) inaccuracies of the sensor itself. So far only disturbances on the measurements due to unmodeled objects in the environment are included. To capture the additional uncertainty, additional measurement noise is added. After taking into account the disturbances by unmodeled objects, unexplainable measurements and sensor failures (Section 3.4), there is no physical reason to expect that the mean value of the true measurements deviates from the expected measurement and that the true measurements are distributed asymmetrically around the mean. Therefore symmetrical noise with mean value zero is added. Two facts justify the modeling of the measurement noise as a normal distribution: (i) the normal distribution maximizes the information entropy among all distributions with known mean and variance, making it the natural choice of underlying distribution for data summarized in terms of sample mean and variance; and (ii) if the underlying phenomena are assumed to have a small, independent effect on the measurement, the central limit theorem states that under certain conditions (such as being independent and identically distributed with finite variance), the sum of a large number of random variables is approximately normally distributed. *If the measurement noise is modeled by a zero mean Gaussian with standard deviation* $\sigma_m$, the conditional probability $P(z \mid x, m, z^\star_{\text{occl}}, k)$ is:

$$P(z \mid x, m, z^\star_{\text{occl}}, k) = \begin{cases} \mathcal{N}(z; z^\star, \sigma_m) & \text{if } k = 0 \\ \mathcal{N}(z; z^\star_{\text{occl}}, \sigma_m) & \text{if } k \geq 1, \end{cases} \tag{9}$$

where the conditional probability $P(z \mid x, m, z^\star_{\text{occl}}, k)$ has two main cases, the first for $k = 0$ where no occlusion is present and the sensor is observing the map $m$, and the second case for $k \geq 1$ where the sensor observes an occluding object. $\sigma_m$ is included in the Bayesian network of Fig. 1.

### 3.3 Marginalization

This section shows the different steps needed to marginalize out the extra state variables in Eq. (1), and motivates the approximation that leads to an analytical sensor model.

The product rule rewrites the sensor model $P(z \mid x, m)$ as:

$$P(z \mid x, m) = \frac{P(z, x, m)}{P(x, m)} = \frac{P(z, x, m)}{P(x)\, P(m)}, \tag{10}$$

since $X$ and $M$ are independent. The numerator is obtained by marginalizing the joint probability of the whole Bayesian network $p_{joint} = P(z, x, m, x_N, n, x_K, k, z^\star_{\text{occl}})$ over $x_N$, $n$, $x_K$, $k$ and $z^\star_{\text{occl}}$:

$$P(z, x, m) = \int_{z^\star_{\text{occl}}} \sum_k \int_{x_K} \sum_n \int_{x_N} p_{joint}\; dx_N\; dx_K\; dz^\star_{\text{occl}}. \tag{11}$$

Using the chain rule to factorize the joint distribution while making use of the conditional dependencies in the Bayesian network (Fig. 1) yields:

$$\begin{aligned} p_{joint} \;=\;& P(z \mid x, m, z^\star_{\text{occl}}, k)\, P(z^\star_{\text{occl}} \mid x_K)\, P(k \mid n, x_N, x, m) \\ & P(x_K \mid x_N, k, x, m)\, P(x_N \mid n)\, P(n)\, P(x)\, P(m)\,. \end{aligned} \tag{12}$$

Substituting (12) and then (11) into (10) gives:

$$\begin{aligned} P(z \mid x, m) \;=\;& \int_{z^\star_{\text{occl}}} \sum_k P(z \mid x, m, z^\star_{\text{occl}}, k) \int_{x_K} P(z^\star_{\text{occl}} \mid x_K) \\ & \sum_n P(k \mid n, x, m)\, P(n)\, P(x_K \mid n, k, x, m)\;\; dx_K\;\; dz^\star_{\text{occl}}, \end{aligned} \tag{13}$$

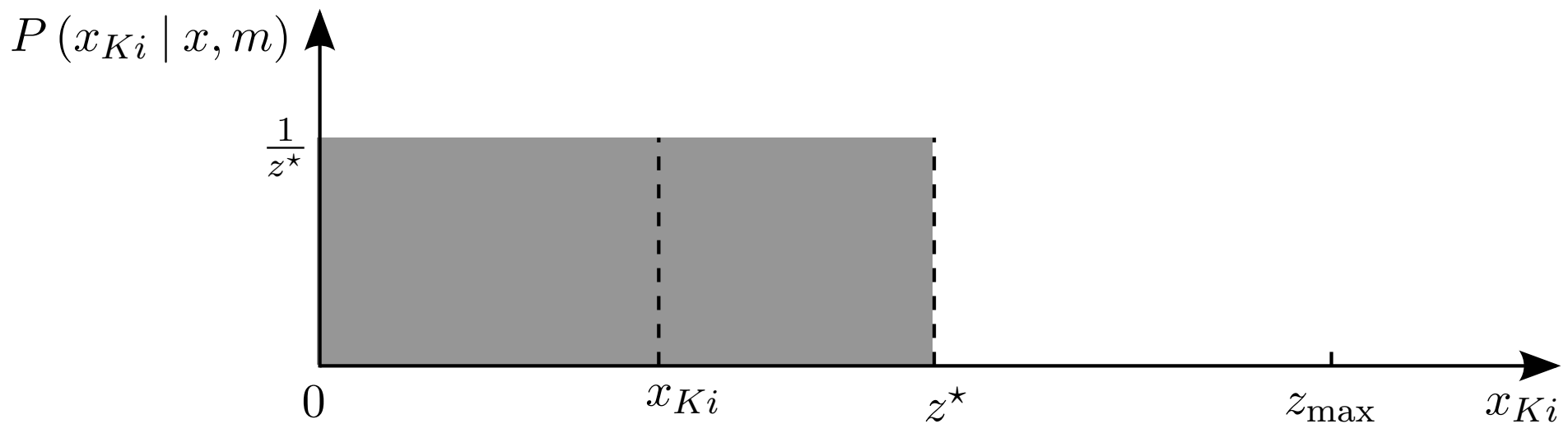

Figure 6: $P\left(x_{Ki} \mid n,k,x,m\right)$ (Eq. (15))

where

$$P\left(x_K \mid n,k,x,m\right) = \int_{x_N} P\left(x_K \mid x_N,k,x,m\right) P\left(x_N \mid n\right) dx_N. \tag{14}$$

Since the binomial distribution $P\left(k \mid n,x_N,x,m\right)$ of Eq. (5) is independent of $x_N$, it is moved out of the integral over $x_N$ in (14), and is further on denoted by $P\left(k \mid n,x,m\right)$.

**Marginalizing $\boldsymbol{x_N}$** Now study the integral over $x_N$ in Eq. (14) and focus on $x_{Nj}$, the position of one unmodeled object. Substituting (3) and (7) results in:

$$\begin{aligned} P\left(x_{Ki} \mid n,k,x,m\right) &= \int_{x_{Nj}=0}^{z^\star} \frac{z_{\max}}{z^\star} \delta\left(x_{Ki} - x_{Nj}\right) \frac{1}{z_{\max}} dx_{Nj} \\ &= \begin{cases} \frac{1}{z^\star} & \text{if } x_{Ki} \leq z^\star \\ 0 & \text{otherwise.} \end{cases} \end{aligned} \tag{15}$$

This equation expresses that $x_{Ki}$ is uniformly distributed when conditioned on $n$, $k$, $x$ and $m$ as shown in Fig. 6. Since all occluding objects are considered independent:

$$P\left(x_K \mid n,k,x,m\right) = \begin{cases} \left(\frac{1}{z^\star}\right)^k & \text{if } \forall\, 0 \leq i \leq k : x_{Ki} \leq z^\star \\ 0 & \text{otherwise.} \end{cases} \tag{16}$$

This equation shows that $P\left(x_K \mid n,k,x,m\right)$ is independent of $n$ and thus can be moved out of the summation over $n$ in Eq. (13):

$$P\left(z \mid x,m\right) = \int_{z^\star_{\text{occl}}} \sum_k P\left(z \mid x,m,z^\star_{\text{occl}},k\right) P\left(z^\star_{\text{occl}} \mid k,x,m\right) P\left(k \mid x,m\right)\ dz^\star_{\text{occl}}, \tag{17}$$

with

$$P\left(k \mid x,m\right) \;=\; \sum_n P\left(k \mid n,x,m\right) P\left(n\right), \tag{18}$$

and

$$P\left(z^\star_{\text{occl}} \mid n,k,x,m\right) \;=\; \int_{x_K} P\left(z^\star_{\text{occl}} \mid x_K\right) P\left(x_K \mid k,x,m\right)\ dx_K. \tag{19}$$

**Marginalizing $n$** First focus on the summation over $n$ in Eq. (18) and substitute (2) and (5):

$$P(k \mid x, m) = \sum_{n=k}^{\infty} \left[ \begin{pmatrix} n \\ k \end{pmatrix} u^k (1-u)^{n-k} (1-p) p^n \right]. \tag{20}$$

Appendix A proves that this infinite sum simplifies to:

$$P(k \mid x, m) = \left(1 - p'\right) p'^k, \tag{21}$$

with

$$p' = \frac{up}{1 - (1-u) p}. \tag{22}$$

**Marginalizing $x_K$** Now focus on the integral over $x_K$ in Eq. (19). Substituting (8) into this equation results in:

$$\begin{aligned} P\left(z^\star_{\text{occl}} \mid k, x, m\right) &= \int_{x_{Kc}} \delta\left(z^\star_{\text{occl}} - x_{Kc}\right) P\left(x_{Kc} \mid k\right) \, dx_{Kc} \\ &= P\left(x_{Kc} = z^\star_{\text{occl}} \mid k, x, m\right). \end{aligned} \tag{23}$$

This equation shows that the conditional probability $P\left(z^\star_{occl} \mid k, x, m\right)$ represents the probability that the perfect measurement of the nearest occluding object is $z^\star_{occl}$, i.e. the probability that the nearest occluding object is located at $z^\star_{\text{occl}}$. This is only the case when one of the $k$ objects along the measurement beam is located such that $z^\star_{occl}$ is measured, while all other objects along the measurement beam are located behind the occluding object, or expressed in probabilities:

$$P\left(z^\star_{\text{occl}} \mid k, x, m\right) = \sum_{i=1}^{k} P\left(x_{K \neq i} \geq z^\star_{\text{occl}} \mid k, x, m\right) P\left(x_{Ki} = z^\star_{\text{occl}} \mid k, x, m\right). \tag{24}$$

Since $x_K$ is uniformly distributed over $[0, z^\star]$ as shown by Eq. (15), it follows that:

$$P\left(x_{Ki} = z^\star_{\text{occl}} \mid k, x, m\right) = \frac{1}{z^\star}, \tag{25}$$

$$P\left(x_{Ki} \geq z^\star_{\text{occl}} \mid k, x, m\right) = \frac{z^\star - z^\star_{\text{occl}}}{z^\star}, \tag{26}$$

and (24) can be written as:

$$P\left(z^\star_{\text{occl}} \mid k, x, m\right) = k \frac{1}{z^\star} \left( \frac{z^\star - z^\star_{occl}}{z^\star} \right)^{k-1}. \tag{27}$$

**Marginalizing $k$** After obtaining expressions for $P(k \mid x, m)$ (Eq. (21)) and $P\left(z^\star_{occl} \mid k, x, m\right)$ (Eq. (27)) we turn the attention to the summation over $k$ in Eq. (17):

$$P\left(z, z^\star_{\text{occl}} \mid x, m\right) = \sum_k P\left(z \mid x, m, z^\star_{\text{occl}}, k\right) P\left(z^\star_{\text{occl}} \mid k, x, m\right) P(k \mid x, m). \tag{28}$$

Split this summation in two parts: one for $k = 0$, when there is no occlusion, and one for $k \geq 1$, and substitute the expressions for $P(k \mid x, m)$ and $P(z \mid x, m, z^\star_{\text{occl}}, k)$ given by Eq. (21) and Eq. (9), respectively:

$$\begin{aligned}
P\left(z, z^\star_{\text{occl}} \mid x, m\right) &= \mathcal{N}\left(z; z^\star, \sigma_m\right) P\left(z^\star_{\text{occl}} \mid k = 0, x, m\right) P\left(k = 0 \mid x, m\right) + \\
&\qquad \mathcal{N}\left(z; z^\star_{\text{occl}}, \sigma_m\right) P\left(z^\star_{\text{occl}} \mid k \geq 1, x, m\right) P\left(k \geq 1 \mid x, m\right) \\
&= \mathcal{N}\left(z; z^\star, \sigma_m\right) P\left(z^\star_{\text{occl}} \mid k = 0, x, m\right) \left(1 - p'\right) + \\
&\qquad \mathcal{N}\left(z; z^\star_{\text{occl}}, \sigma_m\right) \alpha\left(z^\star_{\text{occl}} \mid x, m\right), \qquad (29) \\
\text{where } \alpha\left(z^\star_{\text{occl}} \mid x, m\right) &= P\left(z^\star_{\text{occl}} \mid k \geq 1, x, m\right) P\left(k \geq 1 \mid x, m\right) \\
&= \sum_{k=1}^{\infty} P\left(z^\star_{\text{occl}} \mid k, x, m\right) \left(1 - p'\right) p'^k. \qquad (30)
\end{aligned}$$

Substituting (27) into (30) results in:

$$\alpha\left(z^\star_{\text{occl}} \mid x, m\right) = \sum_{k=1}^{\infty} k \frac{1}{z^\star} \left(\frac{z^\star - z^\star_{occl}}{z^\star}\right)^{k-1} \left(1 - p'\right) p'^k, \qquad (31)$$

which is simplified using Eq. (114) in Appendix A:

$$\begin{aligned}
\alpha\left(z^\star_{\text{occl}} \mid x, m\right) &= \frac{1}{z^\star} \left(1 - p'\right) p' \sum_{k=1}^{\infty} k \left(\frac{z^\star - z^\star_{\text{occl}}}{z^\star} p'\right)^{k-1} \\
&= \frac{p'\left(1 - p'\right)}{z^\star \left[1 - \left(\frac{z^\star - z^\star_{\text{occl}}}{z^\star} p'\right)\right]^2}. \qquad (32)
\end{aligned}$$

Substituting (32) into (29) gives:

$$\begin{aligned}
P\left(z, z^\star_{\text{occl}} \mid x, m\right) = \mathcal{N}\left(z; z^\star, \sigma_m\right) P\left(z^\star_{\text{occl}} \mid k = 0, x, m\right) \left(1 - p'\right) + \\
\mathcal{N}\left(z; z^\star_{\text{occl}}, \sigma_m\right) \frac{\left(1 - p'\right) p'}{z^\star \left[1 - \left(\frac{z^\star - z^\star_{\text{occl}}}{z^\star} p'\right)\right]^2}. \qquad (33)
\end{aligned}$$

**Marginalizing $z^\star_{\text{occl}}$** Substituting (33) into (17) shows that only the integration over $z^\star_{\text{occl}}$ still has to be carried out:

$$P\left(z \mid x, m\right) = (1 - p')\mathcal{N}\left(z; z^\star, \sigma_m\right) + p' \int_{z^\star_{\text{occl}}=0}^{z^\star} \mathcal{N}\left(z; z^\star_{\text{occl}}, \sigma_m\right) \frac{1 - p'}{z^\star \left[1 - \left(\frac{z^\star - z^\star_{occl}}{z^\star} p'\right)\right]^2} dz^\star_{\text{occl}} \, . (34)$$

The first term of the right hand side is a Gaussian distribution around the ideal measurement, multiplied with the probability of no occlusion ($k = 0$). The second term is an integration over all possible positions of the occluding object of a scaled Gaussian distribution centered at the ideal measurement of the occluding object ($z^\star_{\text{occl}}$). The scaling factor represents the probability that the occluding objects are located such that $z^\star_{\text{occl}}$ is measured. From Eq. (20) and Eq. (32) it follows that the scaling factor can be written as:

$$\alpha\left(z^\star_{\text{occl}} \mid x, m\right) = \frac{p\left(1 - p\right)}{z_{\max} \left[1 - \left(1 - \frac{z^\star_{occl}}{z_{\max}}\right) p\right]^2}, \qquad (35)$$

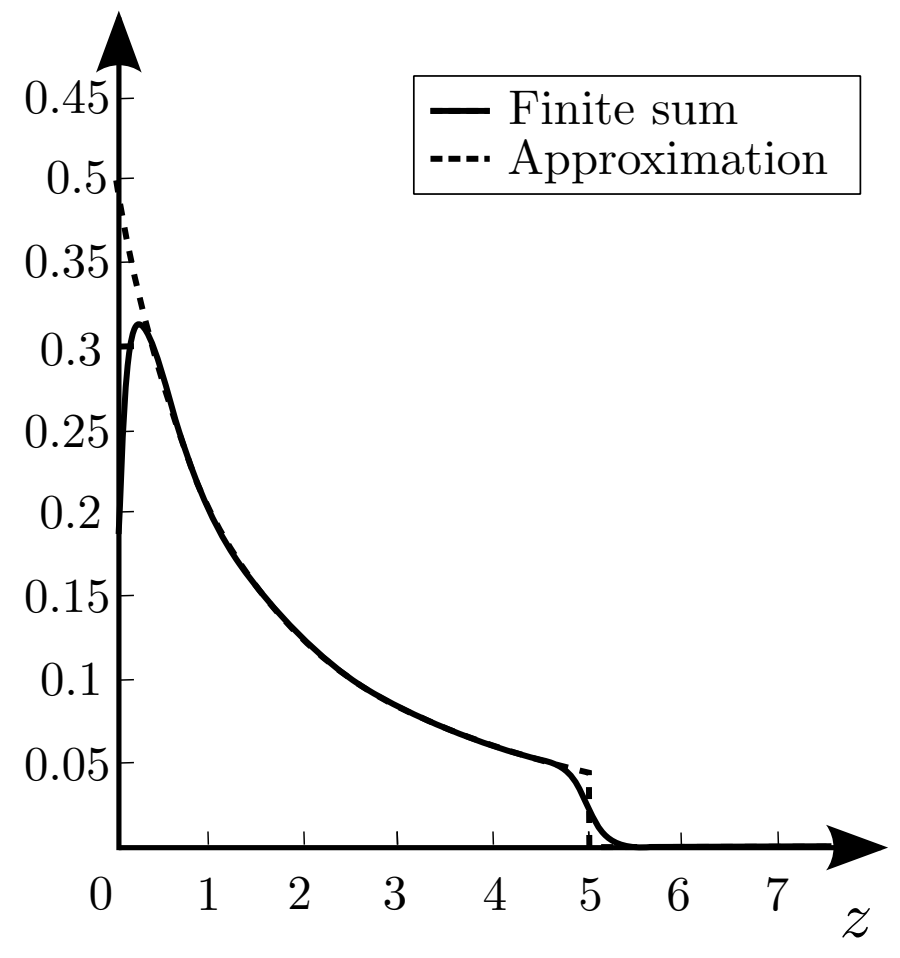

Figure 7: Comparison of the approximation (Eq.(36)) of the integral in Eq. (34) with a finite sum approximation with small step size for $p = 0.8$, $z_{\max} = 10$, $z^{\star} = 5$ and $\sigma_m = 0.15$.

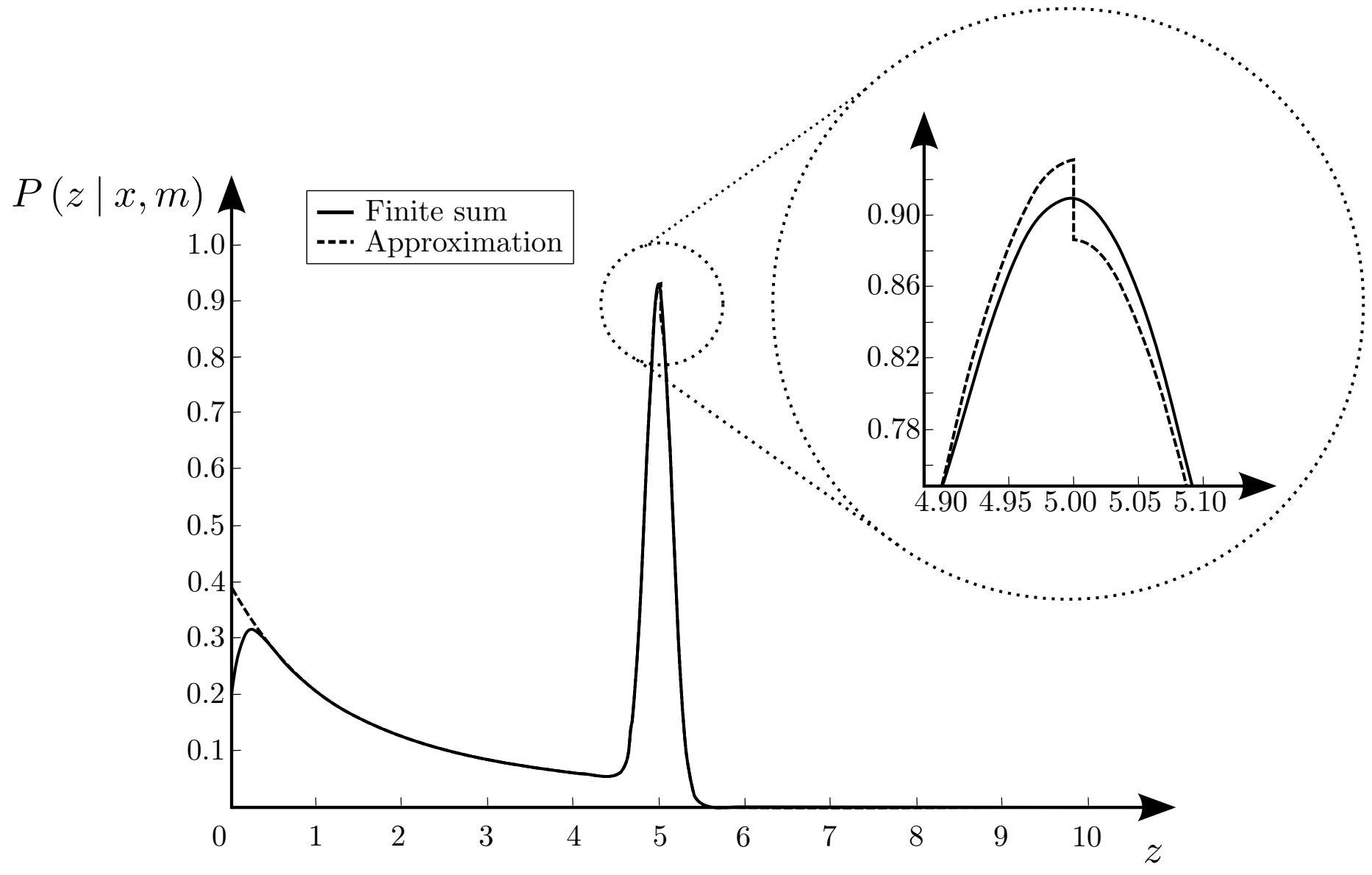

Figure 8: Comparison of the obtained RBBM $P(z \mid x, m)$ (Eq. (37)) with a finite sum approximation of Eq. (34) with small step size for $p = 0.8$, $z_{\max} = 10$, $z^{\star} = 5$ and $\sigma_m = 0.15$.

which is independent of $z^\star$.

Until now, no approximations where made to obtain Eq. (34) for the beam model $P(z \mid x, m)$. The integral over the scaled Gaussian distributions however, cannot be obtained analytically. Therefore, *a first approximation in the marginalization is made* by neglecting the noise on the range measurement in case of occlusion, i.e.: $\mathcal{N}(z; z^\star_{\text{occl}}, \sigma_m) \approx \delta(z - z^\star_{\text{occl}})$. Using this approximation the second term in the right hand side of Eq. (34) becomes:

$$\frac{p'(1-p')}{z^\star\left[1 - \frac{z^\star - z}{z^\star} p'\right]^2} = \frac{p(1-p)}{z_{\max}\left[1 - \left(1 - \frac{z}{z_{\max}}\right) p\right]^2}. \tag{36}$$

Fig. 7 shows the quality of the approximation of the integral in Eq. ( 34) compared to a finite sum approximation with small step size. The approximation introduces a discontinuity around $z = z^\star$. Using the proposed approximation for the integral the resulting beam model is:

$$P(z \mid x, m) \;=\; \begin{cases} (1-p')\,\mathcal{N}(z; z^\star, \sigma_m) + p' \frac{(1-p')}{z^\star\left[1 - \left(\frac{z^\star - z}{z^\star} p'\right)\right]^2} & \text{if } z \le z^\star \\ (1-p')\,\mathcal{N}(z; z^\star, \sigma_m) & \text{otherwise,} \end{cases} \tag{37}$$

as shown in Fig. 8.

The RBBM can be written as a mixture of two components:

$$P(z \mid x, m) = \pi_1 P_{\text{hit}}(z \mid x, m) + \pi_2 P_{\text{occl}}(z \mid x, m), \tag{38}$$

$$\text{with } \pi_1 = \left(1 - p'\right) \tag{39}$$

$$\pi_2 = p' \tag{40}$$

$$P_{\text{hit}}(z \mid x, m) = \mathcal{N}(z; z^\star, \sigma_m) \tag{41}$$

$$P_{\text{occl}}(z \mid x, m) = \begin{cases} \frac{1}{z^\star} \frac{1-p'}{\left[1 - \left(\frac{z^\star - z}{z^\star} p'\right)\right]^2} & \text{if } 0 \le z \le z^\star \\ 0 & \text{otherwise.} \end{cases} \tag{42}$$

### 3.4 Extra Components

Occasionally, range finders produce unexplainable measurements, caused by phantom readings when sonars bounce off walls, or suffer from cross-talk (Thrun et al., 2005). Furthermore additional uncertainty on the measurements is caused by (i) uncertainty in the sensor position, (ii) inaccuracies of the world model and (iii) inaccuracies of the sensor itself. These unexplainable measurements are modeled using a uniform distribution spread over the entire measurement range $[0, z_{\max}]$:

$$P_{\text{rand}}(z \mid x, m) = \begin{cases} \frac{1}{z_{\max}} & \text{if } 0 \le z \le z_{\max}, \\ 0 & \text{otherwise.} \end{cases} \tag{43}$$

Furthermore, sensor failures typically produce max-range measurements, modeled as a point-mass distribution centered around $z_{\max}$:

$$P_{\max}(z \mid x, m) = I(z_{\max}) = \begin{cases} 1 & \text{if } z = z_{\max}, \\ 0 & \text{otherwise.} \end{cases} \tag{44}$$

These two extra components can be added to Eq. (38), resulting in the final RBBM:

$$P(z \mid x, m) = \pi_1 \; P_{\text{hit}}(z \mid x, m) + \pi_2 \; P_{\text{occl}}(z \mid x, m) + \pi_3 \; P_{\text{rand}}(z \mid x, m) + \pi_4 \; P_{\max}(z \mid x, m), \tag{45}$$

where $\pi_3$ and $\pi_4$ are the probabilities that the range finder returns an unexplainable measurement and a maximum reading, respectively. Furthermore,

$$\pi_1 = \left(1 - p'\right)(1 - \pi_3 - \pi_4) \text{ and} \tag{46}$$

$$\pi_2 = p'(1 - \pi_3 - \pi_4), \tag{47}$$

while $P_{\text{hit}}(z \mid x, m)$, $P_{\text{occl}}(z \mid x, m)$, $P_{\text{rand}}(z \mid x, m)$ and $P_{\max}(z \mid x, m)$ are given by (41), (42), (43) and (44) respectively.

### 3.5 Assumptions and Approximations

This section summarizes the assumptions and approximations made to arrive at the RBBM of Eq. (45).
Section 3.2 makes four assumptions:

(i) the probability of the number of unmodeled objects decreases exponentially, Eq. (2);

(ii) the unmodeled object's position is uniformly distributed over the measurement beam (Fig. 3, Eq. (3));

(iii) the positions of the unmodeled objects are independent, Eq. (4); and

(iv) the measurement noise is zero mean normally distributed with standard deviation $\sigma_m$ (Eq. 9).

Furthermore, Section 3.3 makes one approximation to obtain an analytical expression by neglecting the noise on the range measurement in case of occlusion (Eq. (34)).

### 3.6 Interpretation

The following paragraphs give some insights in the RBBM and its derivation.

The mixture representation (45) shows the four possible causes of a range measurement: a hit with the map, a hit with an unmodeled object, an unknown cause resulting in a random measurement and a sensor failure resulting in a maximum reading measurement.

The derivation of Section 3.3 shows that the position of each of the occluding objects is uniformly distributed between the sensor and the ideally measured object in the environment (Eq. (15), Fig. 6). This is perfectly reasonable considering the assumption of uniformly distributed unmodeled objects.

Furthermore, some insights are provided concerning $\alpha(z^\star_{\text{occl}} \mid x, m)$ (Eq. (35), Fig. 7), the probability that the occluding objects are located such that $z^\star_{\text{occl}}$ is measured. First of all, this probability is independent of the location of the ideally measured object in the environment ($z^\star$) (except that this probability is zero for $z > z^\star$). This agrees with intuition, since one expects the measurements caused by the occluding objects to be independent of $z^\star$, the measurement in the case of no occlusion. Second, the probability of sensing unmodeled objects decreases with the range, as expected. This is easily explained with the following thought experiment: if two objects are present with the same likelihood in the perception field of the range finder, and the first object is closest to the range sensor, then the sensor is more likely to measure the first object. To measure the second object, the second object should be present *and* the first object should be absent (Thrun et al., 2005). Moreover, the rate of decrease of the likelihood of sensing unmodeled objects is only dependent on $p$, the degree of appearance of unmodeled objects.

The probability of measuring a feature of the map, and therefore the integral under the scaled Gaussian $(1-p')P_{\text{hit}}(z \mid x, m)$ (45), decreases with the expected range. This is easily explained since the probability that the map is not occluded decreases when the feature is located further away.

Finally, the discontinuity of the RBBM (Fig. 8) was shown to be caused by the only approximation made (Section 3.5). Since the state of the art range sensors are very accurate, neglecting the measurement noise on the measurement of an occluding object is an acceptable approximation. This is also shown by the experiments presented in Section 5.

With respect to the state of the art beam model of Thrun et al. (2005), the model proposed here, Eq. (45), has: (i) a different functional form for the probability of range measurements caused by unmodeled objects, (ii) an intuitive explanation for the discontinuity encountered in the cited paper, and (iii) a reduction in the number of model parameters. Thrun et al. find that $P_{\text{occl}}(z \mid x, m)$ has an exponential distribution. This exponential distribution results from the following underlying assumptions (although not revealed by the authors): (i) the unmodeled objects are equally distributed in the environment and (ii) a beam is reflected with a constant probability at any range. The last assumption equals assuming that the probability that an unmodeled object is located at a certain distance is constant. This assumption fails to capture that the number of unmodeled object is finite and that it is more probable to have a limited number of unmodeled objects than a huge number of them. While we also assume that unmodeled objects are equally distributed in the environment (Eq. (3)), we assume that the number of unmodeled objects is geometrically distributed (Eq. (2)) capturing the finiteness of the number of unmodeled objects and the higher probability of a smaller number of unmodeled objects. The modeling of the finiteness of the number of unmodeled objects and the higher probability of a smaller number of unmodeled objects results in a quadratic decay of $P_{\text{occl}}(z \mid x, m)$, instead of the exponential decay of $P_{\text{occl}}(z \mid x, m)$ found by Thrun et al..

As stated in the previous paragraph, the discontinuity of the RBBM (Fig. 8) is caused by an approximation. While Thrun's model considers the rate of decay of $P_{\text{occl}}(z \mid x, m)$ to be independent of $\pi_2$, the probability of an occlusion, it is shown here that they both depend on the same parameter $p'$ (Eq. (42), Eq. (47)). Therefore the RBBM has fewer parameters than Thrun's model.

### 3.7 Validation

The goal of this section is to show by means of a Monte Carlo simulation[2] that the RBBM, Eq. (45), agrees with the Bayesian network in Fig. 1. A Monte Carlo simulation is an approximate inference method for Bayesian networks. The idea behind the Monte Carlo simulation is to draw random configurations of the network variables $Z$, $X$, $M$, $N$, $X_N = \{X_{Nj}\}_{j=1:n}$, $K$, $X_K = \{X_{Ki}\}_{i=1:k}$ and $Z^{\star}_{\text{occl}}$ and to do this a sufficient number of times. Random configurations are selected by ancestral sampling (Bishop, 2006), i.e. by successively sampling

2. A Monte Carlo simulation is also known as stochastic simulation in the Bayesian network literature (Jensen & Nielsen, 2007).

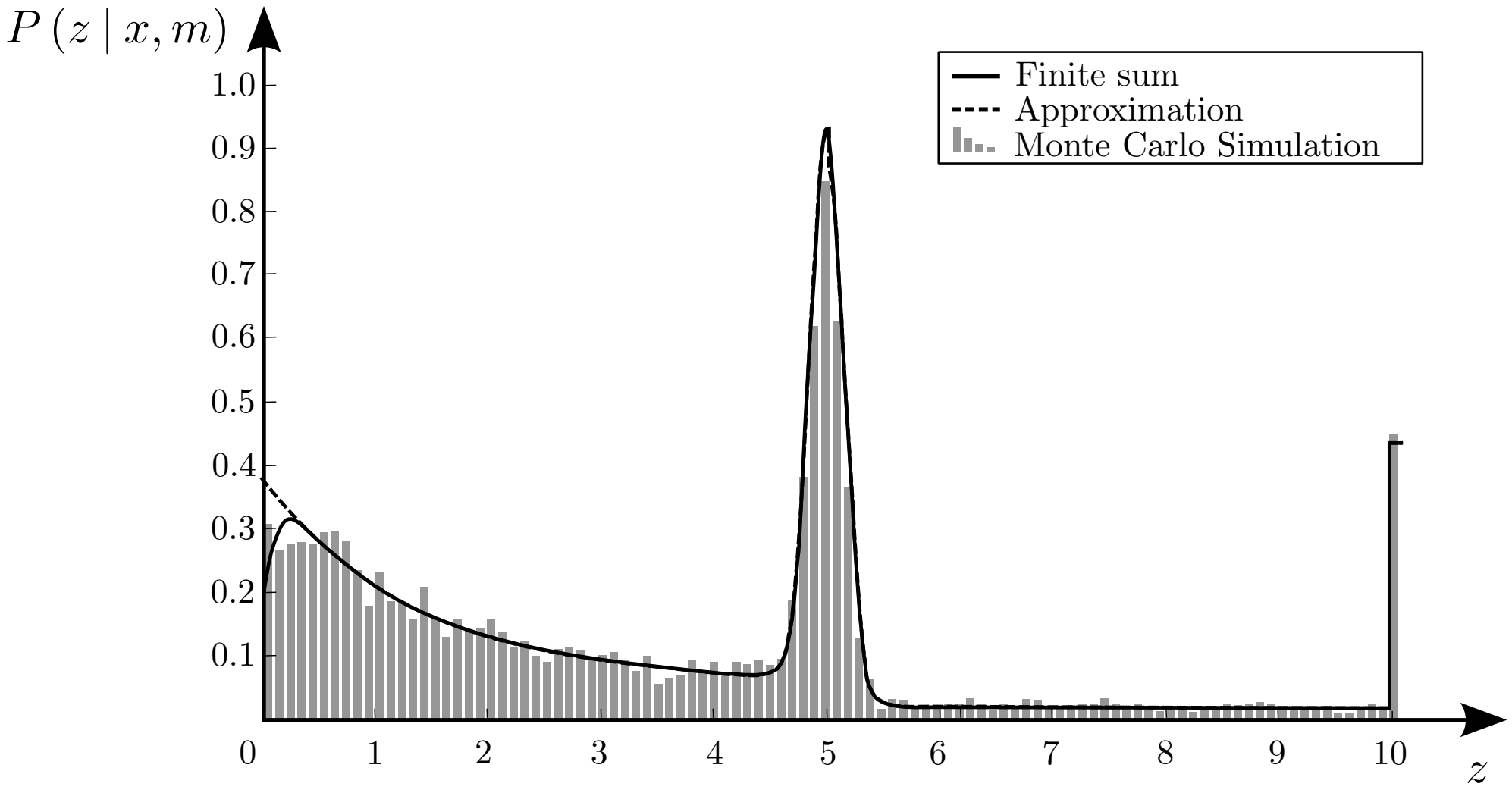

Figure 9: Comparison of the obtained RBBM $P(z \mid x, m)$ (45), a finite sum approximation of Eq. (34) with small stepsize and the normalized histogram of 500 samples obtained by a Monte Carlo Simulation of the proposed Bayesian network (Fig. 1) for $p = 0.8$, $z_{\max} = 10$, $z^{\star} = 5$, $\sigma = 0.15$, $\pi_3 = 0.2$ and $\pi_4 = 0.02$.

the states of the variables following the causal model defined by the directed acyclic graph of the Bayesian network.

Fig. 9 shows that the RBBM agrees with a Monte Carlo simulation with 500 samples of the proposed Bayesian network.

## 4. Variational Bayesian Learning of the Model Parameters

The RBBM, Eq. (45), depends on four independent model parameters:

$$\Theta = \left[\sigma_m, p', \pi_3, \pi_4\right], \tag{48}$$

while $z_{\max}$ is a known sensor characteristic. This set of parameters has a clear physical interpretation; $\sigma_m$ is the standard deviation of the zero mean Gaussian measurement noise in Eq. (9) governing $P_{\text{hit}}(z \mid x, m)$ (Eq. (41)); $p'$, defined in Eq. (21), is the probability that the map is occluded ($P(k \geq 1 \mid x, m)$); $\pi_3$ and $\pi_4$ are the probabilities that the range finder returns an unexplainable measurement (unknown cause) and a maximum reading (sensor failure), respectively.

An alternative non-minimal set of parameters containing all the mixing coefficients $\boldsymbol{\pi} = [\pi_1, \pi_2, \pi_3, \pi_4]$ could be used: $\Theta' = [\sigma_m, \boldsymbol{\pi}]$, provided that the constraint:

$$\sum_{s=1}^{S=4} \pi_s \ = \ 1, \tag{49}$$

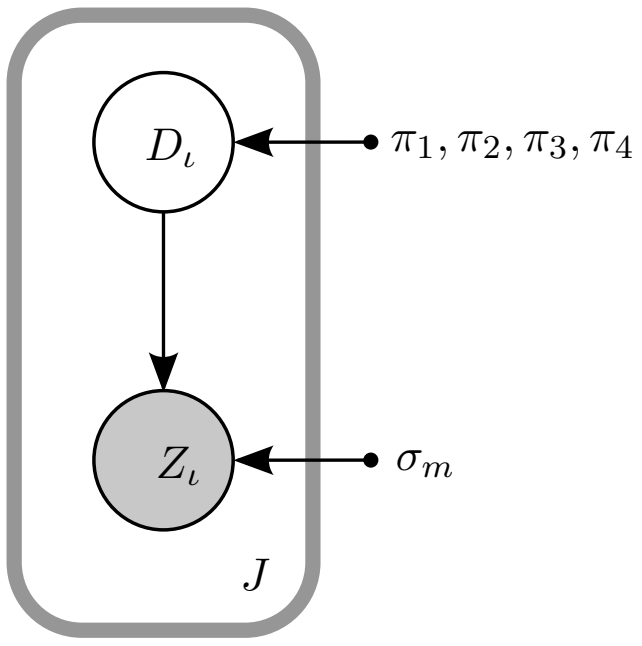

Figure 10: Graphical representation of the mixture measurement model (Eq. (45)) with latent correspondence variable $D_\iota = \{D_{\iota 1}, D_{\iota 2}, D_{\iota 3}, D_{\iota 4}\}$ and model parameters $\Theta' = [\sigma_m, \pi_1, \pi_2, \pi_3, \pi_4]$.

is taken into account. The set of minimal parameters $\Theta$ straightforwardly follows from the non-minimal set $\Theta'$ since:

$$p' = \frac{\pi_2}{1 - \pi_3 - \pi_4}, \tag{50}$$

as can be seen from Eq. (47).

The physical interpretation of the parameters $\Theta$ allows us to initialize them by hand with plausible values. However, another, more flexible way is to learn the model parameters from actual data containing $J$ measurements $\boldsymbol{Z} = \{z_\iota\}_{\iota=1:J}$ with corresponding states $\boldsymbol{X} = \{x_\iota\}_{\iota=1:J}$ and map $m$. Furthermore, learning the model parameters is also a validation for the proposed analytical model: if the learning algorithm succeeds in finding model parameters such that the resulting distribution gives a good explanation of the data, the analytical model is likely to agree well with reality.

In this paper two different estimators[3], a maximum likelihood (ML) (Dempster, Laird, & Rubin, 1977; McLachlan & Krishnan, 1997; Bishop, 2006) and a variational Bayesian (VB) (Beal & Ghahramani, 2003; Bishop, 2006) estimator, are presented to learn the model parameters from data. Section 4.1 derives a maximum likelihood estimator, which is a known approach for this problem, but reformulated for the RBBM. This ML estimator only provides point estimates of the parameters and leads to overfitting since the likelihood function is generally higher for more complex model structures. Therefore, we propose a variational Bayesian (VB) estimator in Section 4.2, which is a new approach for learning the parameters for beam models. The VB estimator is a fully Bayesian learning approach; priors over the unknown parameters are included, complex (overfitting) models are punished, and a full probability distribution over the parameters is obtained.

3. This paper approximately follows the notation by Bishop (2006).

### 4.1 Maximum Likelihood Learning

A maximum likelihood estimator is proposed to identify the model parameters $\Theta$ that maximize the likelihood of the data $\boldsymbol{Z}$ with corresponding $\boldsymbol{X}$ and map $m$:

$$\Theta = \arg\max_{\Theta} \log P\left(\boldsymbol{Z} \mid \boldsymbol{X}, m, \Theta\right). \tag{51}$$

When using the mixture representation of the RBBM (Eq. (45)), the estimation problem can be formulated as finding the ML estimates for the parameters $\Theta' = [\sigma_m, \boldsymbol{\pi}]$ provided that the constraint in Eq. (49) is included. In general it is not known which of the four possible causes actually caused each of the measurements. In that case the ML estimation problem is difficult and lacks a closed-form solution. If however, the corresponding causes of each of the measurements are known, the solution is easily obtained in closed form. Therefore, introduce a latent correspondence variable $\boldsymbol{d} = [d_1, d_2, d_3, d_4]$, representing the unknown cause, using a 1-of-S representation. The elements $d_s$ of $\boldsymbol{d}$ give the probability that the measurement is a result of the $s$'th cause. The graphical representation of the mixture formulation including the latent correspondence $\boldsymbol{d}$ variable is shown in Fig. 10. Although the ML estimation problem lacks a closed-form solution due to the unknown correspondences, an expectation-maximization approach (EM) can solve the problem by iterating an expectation and a maximization step. The expectation step calculates an expectation for the correspondence variables $d_s$ while the maximization step computes the other model parameters under these expectations.

**Algorithm 1** ML estimator for model parameters

**while** convergence criterion not satisfied **do**
  **for all** $z_\iota$ in $\boldsymbol{Z}$, with $\iota = 1 : J$, where $J = |\boldsymbol{Z}|^{-1}$ **do**
    **calculate** $z_m^\star$
    $\eta = \left[\, \pi_1 \, P_{\text{hit}}\left(z_\iota \mid x_\iota, m\right) + \pi_2 \, P_{\text{occl}}\left(z_\iota \mid x_\iota, m\right) + \pi_3 \, P_{\text{rand}}\left(z_\iota \mid x_\iota, m\right) + \pi_4 \, P_{\max}\left(z_\iota \mid x_\iota, m\right)\right]^{-1}$
    $\gamma\left(d_{\iota 1}\right) = \eta \, \pi_1 \, P_{\text{hit}}\left(z_\iota \mid x_\iota, m\right)$
    $\gamma\left(d_{\iota 2}\right) = \eta \, \pi_2 \, P_{\text{occl}}\left(z_\iota \mid x_\iota, m\right)$
    $\gamma\left(d_{\iota 3}\right) = \eta \, \pi_3 \, P_{\text{rand}}\left(z_\iota \mid x_\iota, m\right)$
    $\gamma\left(d_{\iota 4}\right) = \eta \, \pi_4 \, P_{\max}\left(z_\iota \mid x_\iota, m\right)$
  **end for**
  $\pi_1 = J^{-1} \sum_\iota \gamma\left(d_{\iota 1}\right)$
  $\pi_2 = J^{-1} \sum_\iota \gamma\left(d_{\iota 2}\right)$
  $\pi_3 = J^{-1} \sum_\iota \gamma\left(d_{\iota 3}\right)$
  $\pi_4 = J^{-1} \sum_\iota \gamma\left(d_{\iota 4}\right)$
  $p' = \frac{\pi_2}{1-\pi_3-\pi_4}$
  $\sigma_m = \sqrt{\frac{\sum_\iota \gamma(d_{\iota 1})(z_\iota - z_\iota^\star)^2}{\sum_\iota \gamma(d_{\iota 1})}}$
**end while**
**return** $\Theta = [\sigma_m, p', \pi_3, \pi_4]$

The marginal distribution over the correspondence variable $\boldsymbol{d}$ is specified in terms of the mixing coefficients $\pi_s$ such that:

$$P(d_s = 1) = \pi_s, \tag{52}$$

where the parameters $\boldsymbol{\pi}$ must satisfy the following two conditions:

$$\begin{cases} 0 \leq \pi_s \leq 1, \\ \sum_{s=1}^{S} \pi_s = 1. \end{cases} \tag{53}$$

Since $\boldsymbol{d}$ uses a 1-of-S representation, the marginal distribution can be written as:

$$P(\boldsymbol{d}) = \prod_{s=1}^{S} \pi_s^{d_s}. \tag{54}$$

The EM-algorithm expresses the complete-data log likelihood, i.e. the log likelihood of the observed *and* the latent variables:

$$\begin{aligned} \log P\left(\boldsymbol{Z}, \boldsymbol{D} \mid \boldsymbol{X}, \Theta', m\right) \; = \; & \sum_{\iota=1}^{J} \left(d_{\iota 1}\left(\log \pi_1 + \log P_{\text{hit}}\left(z_\iota \mid x_\iota, m\right)\right) + \cdots\right. \\ & d_{\iota 2}\left(\log \pi_2 + \log P_{\text{occl}}\left(z_\iota \mid x_\iota, m\right)\right) + \cdots \\ & d_{\iota 3}\left(\log \pi_3 + \log P_{\text{rand}}\left(z_\iota \mid x_\iota, m\right)\right) + \cdots \\ & \left. d_{\iota 4}\left(\log \pi_4 + \log P_{\max}\left(z_\iota \mid x_\iota, m\right)\right)\right), \end{aligned} \tag{55}$$

where $\boldsymbol{Z} = \{z_\iota\}_{\iota=1:J}$ is the vector containing the observed data and $\boldsymbol{D} = \{\boldsymbol{d}_\iota\}$ is the vector containing the matching correspondences.

**Expectation step:** Taking the expectation of the complete-data log likelihood in Eq. (55) with respect to the posterior distribution of the latent variables gives:

$$\begin{aligned} Q(\Theta', \Theta'^{old}) \; = \; & \mathbb{E}_D\left[\log P\left(\boldsymbol{Z}, \boldsymbol{D} \mid \boldsymbol{X}, \Theta', m\right)\right] \\ = \; & \sum_{\iota=1}^{J} \left(\gamma\left(d_{\iota 1}\right)\left(\log \pi_1 + \log P_{\text{hit}}\left(z_\iota \mid x_\iota, m\right)\right) + \cdots\right. \\ & \gamma\left(d_{\iota 2}\right)\left(\log \pi_2 + \log P_{\text{occl}}\left(z_\iota \mid x_\iota, m\right)\right) + \cdots \\ & \gamma\left(d_{\iota 3}\right)\left(\log \pi_3 + \log P_{\text{rand}}\left(z_\iota \mid x_\iota, m\right)\right) + \cdots \\ & \left.\gamma\left(d_{\iota 4}\right)\left(\log \pi_4 + \log P_{\max}\left(z_\iota \mid x_\iota, m\right)\right)\right), \end{aligned} \tag{56}$$

where $\gamma(d_{\iota s}) = \mathbb{E}[d_{\iota s}]$ is the discrete posterior probability, or responsibility (Bishop, 2006), of cause $s$ for data point $z_\iota$. In the *E-step*, these responsibilities are evaluated using Bayes' theorem, which takes the form:

$$\gamma(d_{\iota 1}) = \mathbb{E}[d_{\iota 1}] = \frac{\pi_1 P_{\text{hit}}(z_\iota \mid x_\iota, m)}{\text{Norm}}, \tag{57}$$

$$\gamma(d_{\iota 2}) = \mathbb{E}[d_{\iota 2}] = \frac{\pi_2 P_{\text{occl}}(z_\iota \mid x_\iota, m)}{\text{Norm}}, \tag{58}$$

$$\gamma(d_{\iota 3}) = \mathbb{E}[d_{\iota 3}] = \frac{\pi_3 P_{\text{rand}}(z_\iota \mid x_\iota, m)}{\text{Norm}}, \text{ and} \tag{59}$$

$$\gamma(d_{\iota 4}) = \mathbb{E}[d_{\iota 4}] = \frac{\pi_4 P_{\max}(z_\iota \mid x_\iota, m)}{\text{Norm}}, \tag{60}$$

where Norm is the normalization constant:

$$\text{Norm} = \pi_1 P_{\text{hit}}(z_\iota \mid x_\iota, m) + \pi_2 P_{\text{occl}}(z_\iota \mid x_\iota, m) + \pi_3 P_{\text{rand}}(z_\iota \mid x_\iota, m) + \pi_4 P_{\text{max}}(z_\iota \mid x_\iota, m). \tag{61}$$

Two measures are derived from the responsibilities:

$$J_s = \sum_{\iota=1}^{J} \gamma(d_{\iota s}), \text{ and} \tag{62}$$

$$\bar{z}_s = \frac{1}{J_s} \sum_{\iota=1}^{J} \gamma(d_{\iota s}) z_\iota, \tag{63}$$

where $J_s$ is the effective number of data points associated with cause $s$, and $\bar{z}_s$ is the mean of the effective data points associated with cause $s$.

**Maximization step:** In the *M-step* the expected complete-data log likelihood in Eq. (56) is maximized with respect to the parameters $\Theta' = [\sigma_m, \boldsymbol{\pi}]$:

$$\Theta'^{new} = \arg\max_{\Theta'} Q(\Theta', \Theta'^{old}). \tag{64}$$

Maximization with respect to $\pi_s$ using a Lagrange multiplier to enforce the constraint $\sum_s \pi_s = 1$ results in:

$$\pi_s = \frac{J_s}{J}, \tag{65}$$

which is the effective fraction of points in the data set explained by cause $s$. Maximization with respect to $\sigma_m$ results in:

$$\sigma_m = \sqrt{\frac{1}{J_1} \sum_{\iota=1}^{J} \gamma_{d_{\iota 1}} (z_\iota - z_\iota^\star)^2}. \tag{66}$$

Algorithm 1 summarizes the equations for the ML estimator, further on called ML-EM algorithm.

### 4.2 Variational Bayesian Learning

The ML estimator only provides point estimates of the parameters and is sensitive to overfitting (Bishop, 2006). Therefore, we propose a variational Bayesian (VB) estimator, which is a new approach for learning the parameters for beam models. The VB estimator is a fully Bayesian learning approach; priors over the unknown parameters are included, complex (overfitting) models are punished, and a full probability distribution over the parameters is obtained. The VB estimator has only a little computational overhead as compared to the ML estimator (Bishop, 2006).

The Bayesian approach attempts to integrate over the possible values of all uncertain quantities rather than optimize them as in the ML approach (Beal, 2003; Beal & Ghahramani, 2003). The quantity that results from integrating out both the latent variables *and* the parameters is known as the marginal likelihood[4]: $P(\boldsymbol{Z}) = \int P(\boldsymbol{Z} \mid \boldsymbol{D}, \Theta)\, P(\boldsymbol{D}, \Theta)\, d(\boldsymbol{D}, \Theta)$, where $P(\boldsymbol{D}, \Theta)$ is a prior over the latent variables and the parameters of the model. Integrating out the parameters penalizes models with more degrees of freedom, since these models can a priori model a larger range of data sets. This property of Bayesian integrations is known as Occam's razor, since it favors simpler explanations for the data over complex ones (Jeffreys & Berger, 1992; Rasmussen & Ghahramani, 2000).

Unfortunately, computing the marginal likelihood, $P(\boldsymbol{Z})$, is intractable for almost all models of interest. The variational Bayesian method constructs a lower bound on the marginal likelihood, and attempts to optimize this bound using an iterative scheme that has intriguing similarities to the standard EM algorithm. To emphasize the similarity with ML-EM, the algorithm based on variational Bayesian inference is further on called VB-EM.

By introducing the distribution $Q$ over the latent variables the complete log marginal likelihood can be decomposed as (Bishop, 2006):

$$\log P(\boldsymbol{Z}) = \mathcal{L}(Q) + KL(Q||P), \tag{67}$$

where

$$\mathcal{L}(Q) = \int Q(\boldsymbol{D}, \mathit{\Theta}) \log \left( \frac{P(\boldsymbol{Z}, \boldsymbol{D}, \mathit{\Theta})}{Q(\boldsymbol{D}, \mathit{\Theta})} \right) d(\boldsymbol{D}, \mathit{\Theta}), \tag{68}$$

and $KL(Q||P)$ is the KL-divergence between $Q$ and $P$. Since this KL-divergence is always greater or equal than zero, $\mathcal{L}(Q)$ is a lower bound on the log marginal likelihood. Maximizing this lower bound with respect to the distribution $Q(\boldsymbol{D}, \Theta)$ is equivalent to minimizing the KL-divergence. If any possible choice for $Q(\boldsymbol{D}, \Theta)$ is allowed, the maximum of the lower bound would occur when the KL-divergence vanishes, i.e. when $Q(\boldsymbol{D}, \Theta)$ is equal to the posterior distribution $P(\boldsymbol{D}, \Theta \mid Z)$. Working with the true posterior distribution is however often intractable in practice. One possible approximate treatment considers only a restricted family of distributions $Q(\boldsymbol{D}, \Theta)$ and seeks the member of this family minimizing the KL-divergence. The variational Bayesian treatment uses a factorized approximation, in this case between the latent variables $\boldsymbol{D}$ and the parameter $\Theta$:

$$Q(\boldsymbol{D}, \Theta) = Q_{\boldsymbol{D}}(\boldsymbol{D})\, Q_{\Theta}(\Theta). \tag{69}$$

The variational approach makes a free form (variational) optimization of $\mathcal{L}(Q)$ with respect to the distributions $Q_{\boldsymbol{D}}(\boldsymbol{D})$ and $Q_{\Theta}(\Theta)$, by optimizing with respect to each of the factors in turn. The general expressions for the optimal factors are (Bishop, 2006):

$$\log Q^{\star}_{\boldsymbol{D}}(\boldsymbol{D}) = \mathbb{E}_{\mathit{\Theta}}[\log P(\boldsymbol{Z}, \boldsymbol{D}, \mathit{\Theta})] + C^{te}, \quad \text{and} \tag{70}$$

$$\log Q^{\star}_{\Theta}(\Theta) = \mathbb{E}_{\boldsymbol{D}}[\log P(\boldsymbol{Z}, \boldsymbol{D}, \mathit{\Theta})] + C^{te}, \tag{71}$$

where $\star$ indicates the optimality. These expressions give no explicit solution for the factors, because the optimal distribution for one of the factors depends on the expectation computed with respect to the other factor. Therefore an iterative procedure, similar to EM, that cycles through the factors and replaces each in turn with a revised optimal estimate is used.

4. To avoid overloading the notation the conditioning on the map $m$ and the positions $\boldsymbol{X} = \{x_\iota\}$ associated with the data $\boldsymbol{Z} = \{z_\iota\}$ is not explicitly written.

**Introducing priors** Since the variational Bayesian approach is a fully Bayesian approach, priors have to be introduced over the parameters $\Theta'' = [\mu, \sigma_m, \boldsymbol{\pi}]$. Remark that in the variational Bayesian estimator not only the standard deviation $\sigma_m$ governing $P_{\text{hit}}(z \mid x, m)$ (Eq. (41)) is estimated but also the means, referred to as $\mu$ further on. Since the analysis is considerably simplified if conjugate prior distributions are used, a Dirichlet prior is chosen for the mixing coefficients $\boldsymbol{\pi}$:

$$P(\boldsymbol{\pi}) = \mathrm{Dir}(\boldsymbol{\pi}|\alpha_0), \tag{72}$$

as well as an independent Gaussian-Wishart prior[5] for the mean $\mu$ and the precision $\lambda_m = \sigma_m^{-1}$ of the Gaussian distribution $P_{\text{hit}}(z \mid x, m)$ (Eq. (41)):

$$P(\mu, \lambda_m) = \mathcal{N}\left(\mu|\bar{\mu}_0, (\beta\lambda_m)^{-1}\right) \mathcal{W}(\lambda_m|W_0, \nu_0). \tag{73}$$

$\alpha_0$ gives the effective prior number of observations associated with each component of the mixture. Therefore, if the value of $\alpha_0$ is set small, the posterior distribution will be mainly influenced by the data rather than by the prior.

**Expectation step** Using these conjugate priors, it can be shown that the factor $Q^{\star}_{\boldsymbol{D}}(\boldsymbol{D})$ can be written as:

$$Q^{\star}_{\boldsymbol{D}}(\boldsymbol{D}) = \prod_{\iota=1}^{J} r_{\iota 1}^{d_{\iota 1}} r_{\iota 2}^{d_{\iota 2}} r_{\iota 3}^{d_{\iota 3}} r_{\iota 4}^{d_{\iota 4}}, \tag{74}$$

where the quantities $r_{\iota s}$ are responsibilities analogous to the $\gamma_{\iota s}$ of Eq. (57) and are given by:

$$r_{\iota s} = \frac{\rho_{\iota s}}{\rho_{\iota 1} + \rho_{\iota 2} + \rho_{\iota 3} + \rho_{\iota 4}}, \tag{75}$$

where

$$\log \rho_{\iota 1} = \mathbb{E}[\log \pi_1] + \mathbb{E}[\log P_{\text{hit}}(z_\iota \mid x_\iota, m)], \tag{76}$$
$$\log \rho_{\iota 2} = \mathbb{E}[\log \pi_2] + \mathbb{E}[\log P_{\text{occl}}(z_\iota \mid x_\iota, m)], \tag{77}$$
$$\log \rho_{\iota 3} = \mathbb{E}[\log \pi_3] + \mathbb{E}[\log P_{\text{rand}}(z_\iota \mid x_\iota, m)], \text{ and} \tag{78}$$
$$\log \rho_{\iota 4} = \mathbb{E}[\log \pi_4] + \mathbb{E}[\log P_{\text{max}}(z_\iota \mid x_\iota, m)]. \tag{79}$$

The above equations can be rewritten as:

$$\log \rho_{\iota 1} = \mathbb{E}[\log \pi_1] + \frac{1}{2}\mathbb{E}[\log |\lambda_m|] - \frac{1}{2}\log(2\pi) - \frac{1}{2}\mathbb{E}_{\mu,\lambda_m}\left[(z_\iota - \mu)^T \lambda (z_\iota - \mu)\right], \tag{80}$$
$$\log \rho_{\iota 2} = \mathbb{E}[\log \pi_2] + \log P_{\text{occl}}(z_\iota \mid x_\iota, m), \tag{81}$$
$$\log \rho_{\iota 3} = \mathbb{E}[\log \pi_3] + \log P_{\text{rand}}(z_\iota \mid x_\iota, m), \text{ and} \tag{82}$$
$$\log \rho_{\iota 4} = \mathbb{E}[\log \pi_4] + \log P_{\text{max}}(z_\iota \mid x_\iota, m), \tag{83}$$

5. The parameters are as defined by Bishop (2006).

in which the expectations can be calculated as follows:

$$\begin{aligned}
\mathbb{E}\left[\log \pi_s\right] &= \Psi\left(\alpha_s\right) - \Psi\left(\alpha_1 + \alpha_2 + \alpha_3 + \alpha_4\right), && (84)\\
\mathbb{E}\left[\log \left|\lambda_m\right|\right] &= \Psi\left(\frac{\nu}{2}\right) + \log 2 + \log |W|, \text{ and} && (85)\\
\mathbb{E}_{\mu,\lambda_m}\left[\left(z_\iota - \mu\right)^T \lambda \left(z_\iota - \mu\right)\right] &= \beta^{-1} + \nu\left(z_\iota - \bar{\mu}\right)^T W \left(z_\iota - \bar{\mu}\right), && (86)
\end{aligned}$$

where $\Psi$ is the digamma function.

Three measures are derived from the responsibilities:

$$\begin{aligned}
J_s &= \sum_{\iota=1}^{J} r_{\iota s}, && (87)\\
\bar{z}_s &= \frac{1}{J_s}\sum_{\iota=1}^{J} r_{\iota s} z_\iota, \text{ and} && (88)\\
C_s &= \frac{1}{J_s}\sum_{\iota=1}^{J} r_{\iota s}\left(z_\iota - \bar{z}_s\right)\left(z_\iota - \bar{z}_s\right)^T, && (89)
\end{aligned}$$

where $J_s$ is the effective number of data points associated with cause $s$, $\bar{z}_s$ is the mean of the effective data points associated with cause $s$ and $C_s$ the covariance of the effective data points associated with cause $s$. Due to the similarity with the E-step of the EM-algorithm, the step of calculating the responsibilities in the variational Bayesian inference is known as the variational E-step.

**Maximization step** In accordance with the graphical representation in Fig. 10, it can be shown that the variational posterior $Q_\Theta(\Theta)$ factorizes as $Q(\boldsymbol{\pi})\, Q(\mu_1, \sigma_m)$ and that the first optimal factor is given by a Dirichlet distribution:

$$Q^\star(\boldsymbol{\pi}) = \mathrm{Dir}\left(\boldsymbol{\pi}|\boldsymbol{\alpha}\right), \tag{90}$$

with

$$\alpha_s = \alpha_0 + J_s. \tag{91}$$

The second optimal factor is given by a Gaussian-Wishart distribution:

$$Q^\star\left(\mu_1, \lambda_m\right) = \mathcal{N}\left(\mu_1|\bar{\mu}, \left(\beta\lambda_m\right)^{-1}\right)\mathcal{W}\left(\lambda_m|W, \nu\right), \tag{92}$$

with

$$\begin{aligned}
\beta &= \beta_0 + J_1, && (93)\\
\bar{\mu} &= \frac{1}{\beta}\left(\beta_0\bar{\mu}_0 + J_1\bar{z}_1\right), && (94)\\
W^{-1} &= W_0^{-1} + J_1 C_1 + \frac{\beta_0 J_1}{\beta_0 + J_1}\left(\bar{z}_1 - \bar{\mu}_0\right)\left(\bar{z}_1 - \bar{\mu}_0\right)^T, \text{ and} && (95)\\
\nu &= \nu_0 + J_1. && (96)
\end{aligned}$$

These update equations are analogous to the M-step of the EM-algorithm for the maximum likelihood solution and is therefore known as the variational M-step. The variational M-step computes a distribution over the parameters (in the conjugate family) rather than a point estimate as in the case of the maximum likelihood estimator. This distribution over the parameters allows us to calculate a predictive density $P(z \mid Z)$.

Due to the use of conjugate priors, the integrals in the predictive density can be calculated analytically:

$$P(z \mid \boldsymbol{Z}) = \frac{\alpha_1 \mathrm{St}\left(z|\bar{\mu}, \frac{\nu\beta}{1+\beta}W, \nu\right) + \alpha_2 P_{\mathrm{occl}}(z \mid x, m) + \alpha_3 P_{\mathrm{rand}}(z \mid x, m) + \alpha_4 P_{\max}(z \mid x, m)}{\alpha_1 + \alpha_2 + \alpha_3}, \tag{97}$$

where St (.) is a Student's t-distribution. When the size $J$ of the data is large, the Student's t-distribution approximates a Gaussian and the predictive distribution can be rewritten as:

$$P(z \mid \boldsymbol{Z}) = \frac{\alpha_1 \mathcal{N}(z|\mu, \lambda_m) + \alpha_2 P_{\mathrm{occl}}(z \mid x, m) + \alpha_3 P_{\mathrm{rand}}(z \mid x, m) + \alpha_4 P_{\max}(z \mid x, m)}{\alpha_1 + \alpha_2 + \alpha_3 + \alpha_4}. \tag{98}$$

If point estimates are desired for the parameters, maximum a posteriori estimates can be obtained as follows:

$$\hat{\pi}_s = \mathbb{E}[\pi_s] = \frac{\alpha_s}{\alpha_1 + \alpha_2 + \alpha_3 + \alpha_4} \tag{99}$$

$$\sigma_m = \left(\frac{\nu\beta}{1+\beta}W\right)^{-\frac{1}{2}}, \text{ and} \tag{100}$$

$$p' = \frac{\hat{\pi_2}}{1 - \hat{\pi_3} - \hat{\pi_4}}. \tag{101}$$

Algorithm 2 summarizes the equations for the VB-EM estimator.

## 5. Experiments

The goal of this section is threefold: (i) to learn the model parameters (Eq. (48)) of the RBBM (Eq. (45)) from experimental data, (ii) to compare the results of the proposed ML-EM and VB-EM estimator (Section 4), and (iii) to compare the results of the proposed estimators with the learning approach of Thrun's model proposed by Thrun et al. (2005). To this end two experimental setups from different application areas in robotics are used. The data for the first learning experiment is gathered during a typical mobile robot application in which the robot is equipped with a laser scanner and is travelling in an office environment. The data for the second learning experiment is gathered during a typical industrial pick-and-place operation in a human populated environment. A laser scanner is mounted on the industrial robot to make it aware of people and other unexpected objects in the robot's workspace.

To see how well the learned model explains the experiment, the learned continuous pdf $P(z \mid x, m, \Theta)$ of Eq. (45) has to be compared with the discrete pdf of the experimental data (histogram) $H(z)$. To this end, the learned pdf is first discretized using the same bins $\{z_f\}_{f=1:F}$ as the experimental pdf. To quantize the difference between the learned and the

---

**Algorithm 2** Variational Bayesian estimator for model parameters

---

**while** convergence criterion not satisfied
  **for all** $z_\iota$ in $\boldsymbol{Z}$, with $\iota = 1 : J$, where $J = |\boldsymbol{Z}|^{-1}$
    **calculate** $z_m^\star$
    $\rho_{\iota 1} = \exp\left[\Psi\left(\alpha_1\right) - \Psi\left(\alpha_1 + \alpha_2 + \alpha_3 + \alpha_4\right) + \frac{1}{2}\left(\Psi\left(\frac{\nu}{2}\right) + \log 2 + \log|W|\right) - \frac{1}{2}\log\left(2\boldsymbol{\pi}\right)\ldots\right.$
      $\left.\cdots - \frac{1}{2}\left(\beta^{-1} + \nu\left(z_\iota - \bar{\mu}\right)^T W\left(z_\iota - \bar{\mu}\right)\right)\right]$
    $\rho_{\iota 2} = \exp\left[\Psi\left(\alpha_2\right) - \Psi\left(\alpha_1 + \alpha_2 + \alpha_3 + \alpha_4\right) + \log P_{\text{occl}}\left(z_\iota \mid x, m\right)\right]$
    $\rho_{\iota 3} = \exp\left[\Psi\left(\alpha_3\right) - \Psi\left(\alpha_1 + \alpha_2 + \alpha_3 + \alpha_4\right) + \log P_{\text{rand}}\left(z_\iota \mid x, m\right)\right]$
    $\rho_{\iota 4} = \exp\left[\Psi\left(\alpha_4\right) - \Psi\left(\alpha_1 + \alpha_2 + \alpha_3 + \alpha_4\right) + \log P_{\max}\left(z_\iota \mid x, m\right)\right]$
    $\eta = \rho_{\iota 1} + \rho_{\iota 2} + \rho_{\iota 3} + \rho_{\iota 4}$
    $r_{\iota 1} = \eta^{-1}\rho_{\iota 1}$
    $r_{\iota 2} = \eta^{-1}\rho_{\iota 2}$
    $r_{\iota 3} = \eta^{-1}\rho_{\iota 3}$
    $r_{\iota 4} = \eta^{-1}\rho_{\iota 4}$
  **end for**
  $J_1 = \sum_{\iota=1}^{J} r_{\iota 1}$
  $J_2 = \sum_{\iota=1}^{J} r_{\iota 2}$
  $J_3 = \sum_{\iota=1}^{J} r_{\iota 3}$
  $J_4 = \sum_{\iota=1}^{J} r_{\iota 4}$
  $\bar{z}_1 = \frac{1}{J_1}\sum_{j=1}^{J} r_{\iota 1} z_\iota$
  $C_1 = \frac{1}{J_1}\sum_{\iota=1}^{J} r_{\iota 1}\left(z_\iota - \bar{z}_1\right)\left(z_\iota - \bar{z}_1\right)^T,$
  $\alpha_1 = \alpha_0 + J_1.$
  $\alpha_2 = \alpha_0 + J_2.$
  $\alpha_3 = \alpha_0 + J_3.$
  $\alpha_4 = \alpha_0 + J_4.$
  $\beta = \beta_0 + J_1$
  $\bar{\mu} = \frac{1}{\beta}\left(\beta_0\bar{\mu}_0 + J_1\bar{z}_1\right)$
  $W^{-1} = W_0^{-1} + J_1 C_1 + \frac{\beta_0 J_1}{\beta_0 + J_1}\left(\bar{z}_1 - \bar{\mu}_0\right)\left(\bar{z}_1 - \bar{\mu}_0\right)^T$
  $\nu = \nu_0 + J_1$

  $\pi_1 = \frac{\alpha_1}{\alpha_1+\alpha_2+\alpha_3+\alpha_4}$
  $\pi_2 = \frac{\alpha_2}{\alpha_1+\alpha_2+\alpha_3+\alpha_4}$
  $\pi_3 = \frac{\alpha_3}{\alpha_1+\alpha_2+\alpha_3+\alpha_4}$
  $\pi_4 = \frac{\alpha_4}{\alpha_1+\alpha_2+\alpha_3+\alpha_4}$
  $p' = \frac{\pi_2}{1-\pi_3-\pi_4}$
  $\sigma_m = \left(\frac{\nu\beta}{1+\beta}W\right)^{-\frac{1}{2}}$
**end while**
**return** $\{\alpha_1, \alpha_2, \alpha_3, \alpha_4, \beta, \bar{\mu}, W, \nu, \Theta'' = [\mu, \sigma_m, p', \pi_3, \pi_4]\}$

---

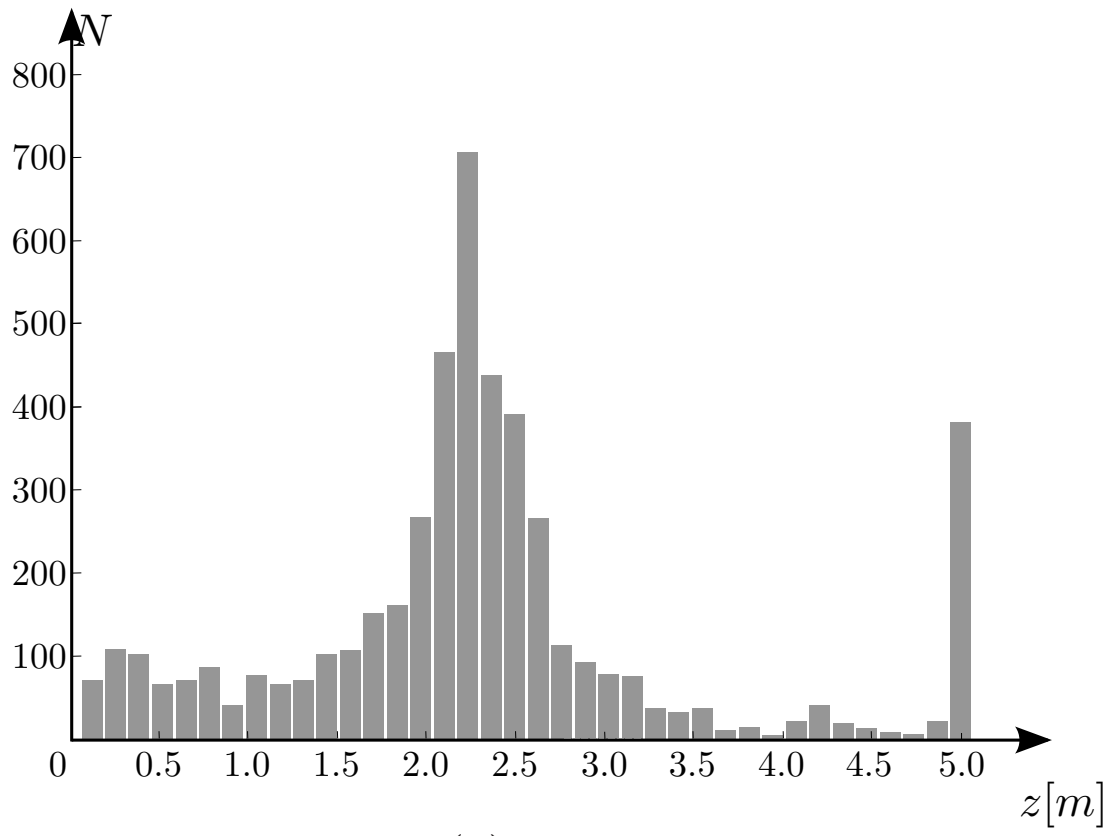

(a) Short range

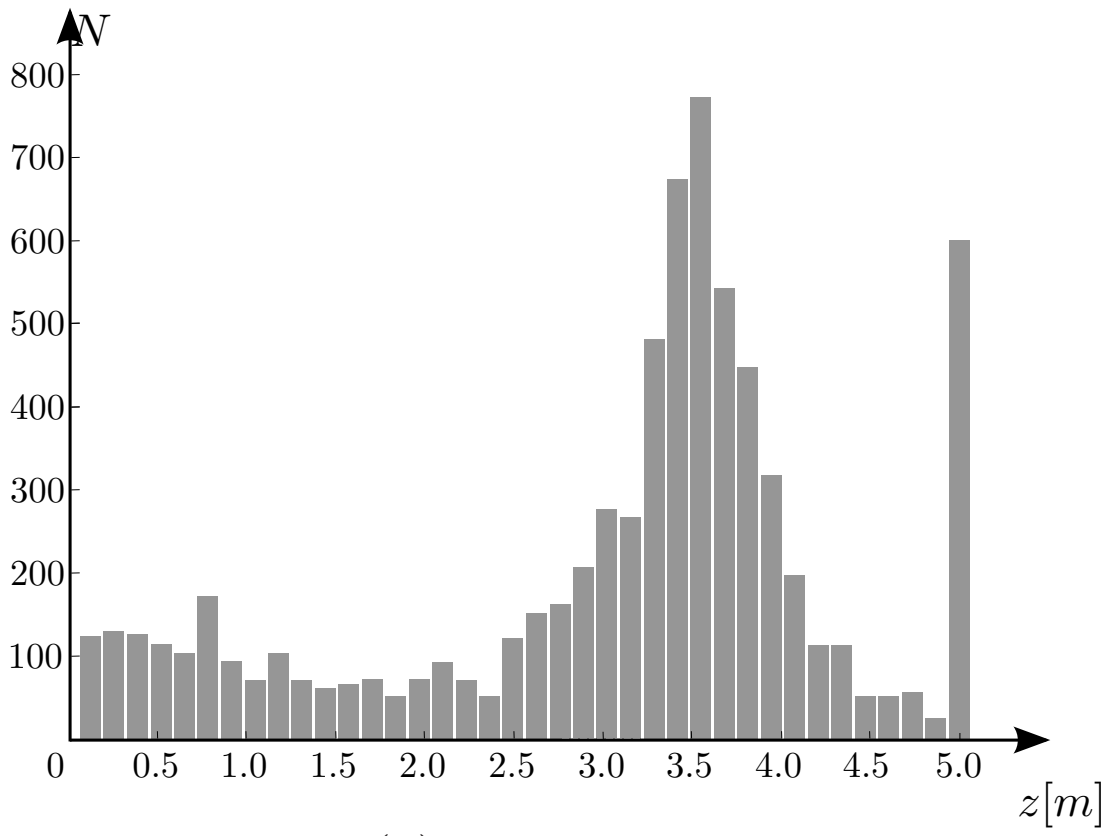

(b) Long range

Figure 11: Data for second learning experiment reported by Thrun et al. (2005). These data consist of two series of measurements obtained with a mobile robot traveling through a typical office environment. From the set of measurements 10000 measurements that are centered around two different expected ranges are selected.

experimental pdf two 'distance' measures are used: the discrete KL-divergence:

$$d_1 = KL\left(H||P\right) \approx \sum_{f=1}^{F} H\left(z_f\right) \log \frac{H\left(z_f\right)}{P\left(z_f \mid x, m, \Theta\right)}, \tag{102}$$

and the square root of the discrete Hellinger distance:

$$d_2 = DH\left(H||P\right) \approx \sqrt{\sum_{f=1}^{F} \left(H\left(z_f\right)^{\frac{1}{2}} - P\left(z_f \mid x, m, \Theta\right)^{\frac{1}{2}}\right)^2}. \tag{103}$$

The latter is known to be a valid symmetric distance metric (Bishop, 2006).

### 5.1 First Learning Experiment

In a first learning experiment, the experimental data reported by Thrun et al. (2005) is used. The data consists of two series of measurements obtained with a mobile robot traveling through a typical office environment. From the set of measurements, 10000 measurements that are centered around two different expected ranges, are selected. The two obtained sets with different expected ranges are shown in Fig. 11. The parameters of the learning algorithms are listed in Table 1. Fig. 12 and Table 2 show the results of the ML-EM and VB-EM estimators for the RBBM compared to the results of the ML estimator for Thrun's model (Thrun et al., 2005) for these two sets. The results are obtained by running the learning algorithms for 30 iteration steps.

| ML-EM RBBM | VB-EM RBBM | | ML-EM Thrun's model |
|---|---|---|---|
| $\sigma_{m,init} = 0.5$ | $p'_{init} = \frac{1}{3}$ | $\alpha_{3,init} = \frac{1}{8}$ | $\sigma_{m,init} = 0.5$ |
| $p'_{init} = 0.4$ | $\beta_{init} = 5000$ | $\alpha_{4,init} = \frac{1}{8}$ | $z_{hit,init} = 0.4$ |
| $\pi_{3,init} = 0.2$ | $W_{init} = 12$ | $\beta_0 = 5$ | $z_{short,init} = 0.3$ |
| $\pi_{4,init} = 0.1$ | $\bar{\mu}_{init} = x_{mp}$ | $W_0 = 50$ | $z_{max,init} = 0.1$ |
| | $\nu_{init} = 100$ | $\bar{\mu}_0 = x_{mp}$ | $z_{rand,init} = 0.2$ |
| | $\alpha_{1,init} = \frac{5}{8}$ | $\nu_0 = 100$ | $\lambda_{short,init} = 0.1$ |
| | $\alpha_{2,init} = \frac{1}{8}$ | $\alpha_0 = 1$ | |

Table 1: EM-parameters for first and second learning experiment (all in SI-units). In the ML approaches, the mean of $P_{\text{hit}}(z \mid x, m)$ is set to $x_{mp}$, i.e. the most probable bin of the histogram of the training set $H(z)$.

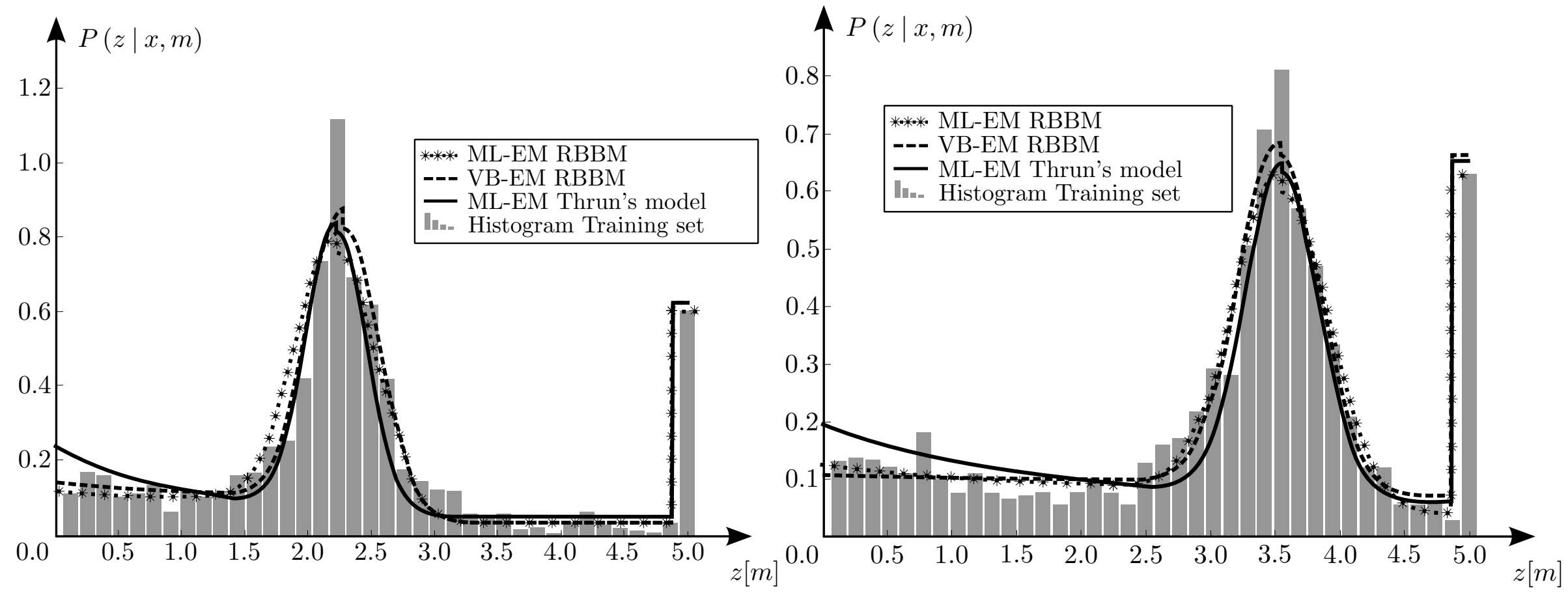

(a) Short range

(b) Long range

Figure 12: Comparison of the results of the ML-EM and VB-EM estimators for the RBBM and the results of a maximum likelihood estimator for Thrun's model (Thrun et al., 2005) for the data of Fig. 11.

The proposed ML-EM and VB-EM estimator outperform the ML-EM estimator for Thrun's model for the studied data sets. Despite the reduced number of parameters of the RBBM compared to Thrun's model (Section 3.6), the RBBM has at least the same representational power.

| Experiment | $\boldsymbol{d_1}$ (Eq. (102)) | | | $\boldsymbol{d_2}$ (Eq. (103)) | | |
|---|---|---|---|---|---|---|
| | ML-EM RBBM | VB-EM RBBM | ML-EM Thrun's model | ML-EM RBBM | VB-EM RBBM | ML-EM Thrun's model |
| short range | 0.5295 | 0.5127 | 0.7079 | 0.3166 | 0.2971 | 0.5629 |
| long range | 0.4366 | 0.4368 | 0.5852 | 0.1683 | 0.2100 | 0.3481 |
| average | 0.4830 | 0.4747 | 0.6465 | 0.2425 | 0.2535 | 0.4555 |

Table 2: Discrete KL-divergence ($d_1$) and square root Hellinger distance ($d_2$) for the first learning experiment between the training set and the results of the ML-EM and VB-EM estimators for the RBBM and the ML-EM estimator for Thrun's model (Thrun et al., 2005).

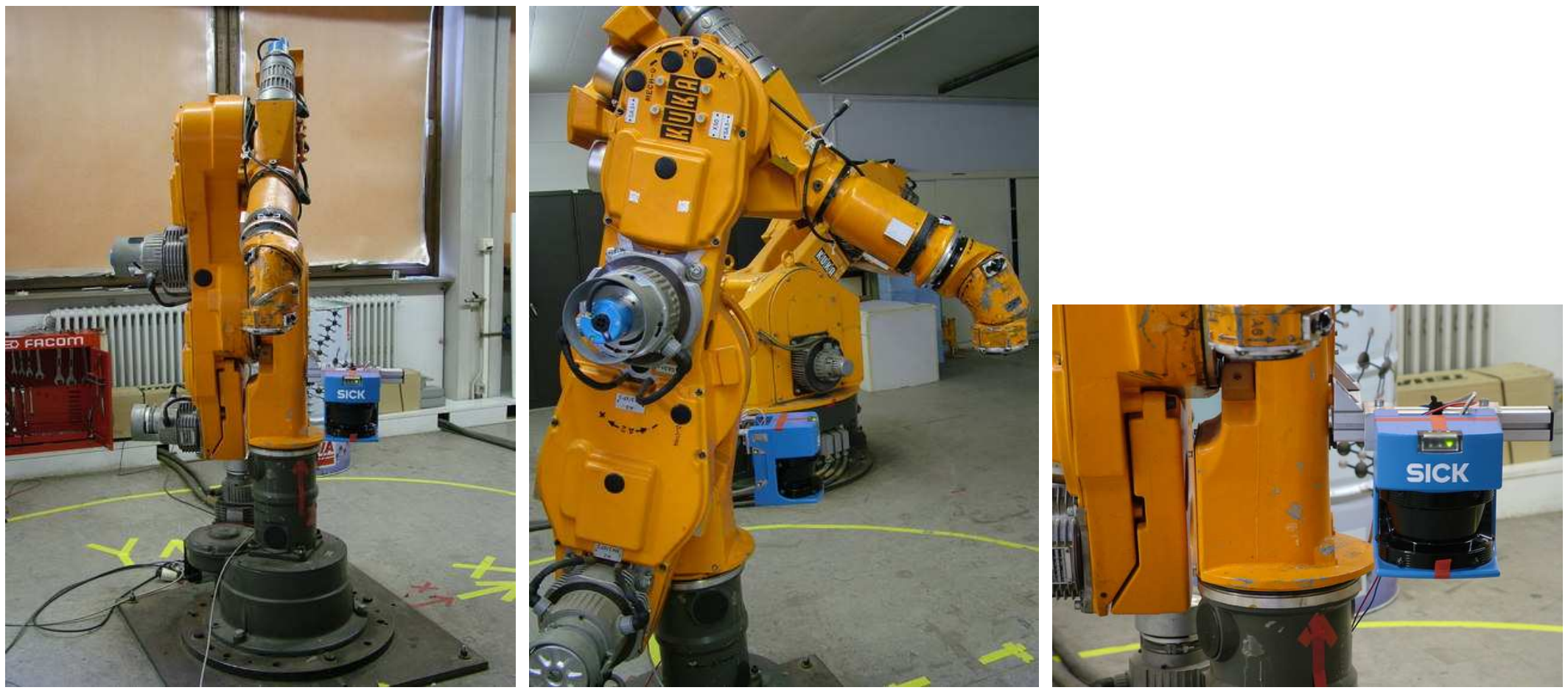

(a) Front view (b) Side view (c) Zoomed front view

Figure 13: Setup for the second learning experiment with a Sick LMS 200 laser scanner mounted on the first axis of an industrial Kuka 361 robot.

### 5.2 Second Learning Experiment

The data for the second learning experiment is gathered during the execution of a typical industrial pick-and-place operation in a human-populated environment. A Sick LMS 200 laser scanner is mounted on the first axis of an industrial Kuka 361 robot (Fig. 13). The laser scanner provides measurements of the robot environment and therefore of people and other unexpected objects in the robot's workspace. Processing these measurements is a first step towards making industrial robots aware of their possibly changing environment and as such of moving these robots out of their cages.

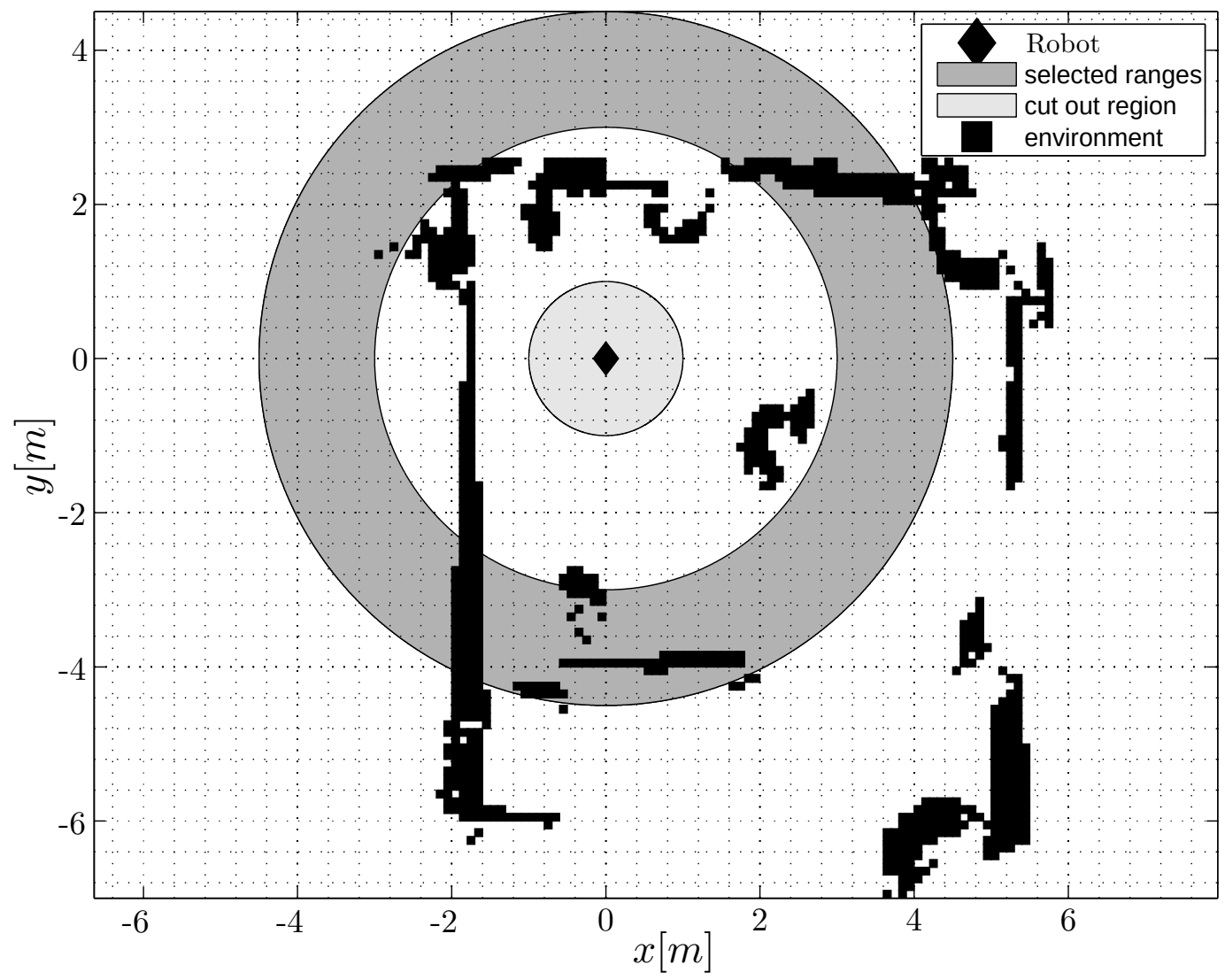

Figure 14: Map build of the robot's static environment, i.e. without any unexpected objects or people moving around, by rotating the first axis of the industrial robot. For safety reasons, people were not allowed to move inside the safety region (circle with radius 1$m$). Therefore, measurements smaller than 1m are discarded. The studied expected ranges in the second learning experiment range from 3.0m to 4.5m in steps of 0.1m and are indicated in the figure by the selected ranges region.

In a first step, a map (Fig. 14) is build of the robot's static environment, i.e. without any unexpected objects or people moving around, by rotating the first axis of the industrial robot. Next, the robot performs a pick-and-place operation while a number of people are walking around at random in the robot environment. Different sets of measurements are acquired each with a different number of people. Similar to the first learning experiment, measurements are selected centered around different expected ranges from the acquired data. The studied expected ranges in the second learning experiment range from 3.0m to 4.5m in steps of 0.1m (Fig. 14). For safety reasons, people were not allowed to move closer than 1m to the robot, i.e. the safety region (Fig. 14). Therefore, measurements smaller than 1m are discarded.

From the data, the model parameters are learning using the same learning parameters as in the first learning experiment (Table. 1).

Table 3 shows the Kullback Leibler divergence (Eq. (102)) and the Hellinger distance (Eq. (103)) averaged for the studied expected range for the different set of measurements after running the ML-EM and VB-EM estimators for the RBBM and the ML estimator for Thrun's model (Thrun et al., 2005). The results were obtained after running each of the learning algorithms for 30 iteration steps.

| Experiment | $\boldsymbol{d_1}$ (Eq. (102)) | | | $\boldsymbol{d_2}$ (Eq. (103)) | | |
|---|---|---|---|---|---|---|
| number of people | ML-EM RBBM | VB-EM RBBM | ML-EM Thrun's model | ML-EM RBBM | VB-EM RBBM | ML-EM Thrun's model |
| 1 | 1.7911 | 1.5271 | 1.9697 | 5.6141 | 4.3449 | 6.5582 |
| 2 | 1.8002 | 1.5172 | 1.9735 | 5.7038 | 4.3334 | 6.6119 |
| 3 | 1.7789 | 1.5199 | 1.9606 | 5.6033 | 4.3468 | 6.5365 |
| 4 | 1.8277 | 1.5140 | 1.9853 | 5.7563 | 4.2972 | 6.6744 |
| 6 | 1.8007 | 1.5168 | 1.9655 | 5.6483 | 4.3126 | 6.5596 |
| 8 | 1.7676 | 1.5157 | 1.9498 | 5.4989 | 4.2843 | 6.4257 |
| average | 1.7944 | 1.5185 | 1.9674 | 5.6375 | 4.3199 | 6.5611 |

Table 3: Discrete KL-divergence ($d_1$) and square root Hellinger distance ($d_2$) averaged for the studied expected range for the different set of measurements of the second learning experiment. Distances are between the training set and the results of the ML-EM and VB-EM estimators for the RBBM and the ML-EM estimator for Thrun's model (Thrun et al., 2005). The first column indicates the number of people walking around in the environment in that particular set of measurements.

The proposed ML-EM and VB-EM estimator outperform the ML-EM estimator for Thrun's model for the studied data sets. Despite the reduced number of parameters of the RBBM compared to Thrun's model (Section 3.6), the RBBM has at least the same representational power.

## 6. Adaptive Full Scan Model

This section extends the RBBM to an adaptive full scan model for dynamic environments; adaptive, since it automatically adapts to the local density of samples when using sample-based representations; full scan, since the model takes into account the dependencies between individual beams.

In many applications using a range finder, the posterior is approximated by a finite set of samples (histogram filter, particle filters). The peaked likelihood function associated with a range finder (small $\sigma_m$ due to its accuracy) is problematic when using such finite set of samples. The likelihood $P(z \mid x, m)$ is evaluated at all samples, which are approximately distributed according to the posterior estimate. Basic sensor models typically assume that the estimate $x$ and the map $m$ are known exactly, that is, they assume that one of the samples corresponds to the true value. This assumption, however, is only valid in the limit of infinitely many samples. Otherwise, the probability that a value exactly corresponds to the true location is virtually zero. As a consequence, these peaked likelihood functions do not adequately model the uncertainty due to the finite, sample-based representation of the posterior (Pfaff et al., 2007). Furthermore, the use of a basic range finder model typically results in even more peaked likelihood models, especially when using a large number of beams per measurement, due to multiplication of probabilities. In practice, the problem of peaked likelihoods, is dealt with in various ways: sub-sampling the measurement (fewer beams); introducing minimal likelihoods for beams; inflating the measurement uncertainty; or other means of regularization of the resulting likelihoods. These solutions are not satisfactory however, since the additional uncertainty due to the sample-based representation is not known in advance. The additional uncertainty strongly varies with the number of samples and the uncertainty of the estimate (Pfaff et al., 2006). Fox (2003) proposes to dynamically adapt the number of samples by means of KLD sampling (KLD stands for Kullback-Leibler divergence). For very peaked likelihoods however, this might result in a huge number of samples. Lenser and Veloso (2000) and Thrun, Fox, Burgard, and Dellaert (2001) ensure that a critical mass of samples is located at the important parts of the state space by sampling from the observation model. Sampling from the observation model however, is often only possible in an approximate and inaccurate way. Pfaff et al. (2006) introduced an *adaptive* beam model for dynamic environments, which explicitly takes location uncertainty due to the sample-based representation into account. They compute the additional uncertainty due to the sample-based representation, using techniques from density estimation. When evaluating the likelihood function at a sample, they consider a certain region around the sample, depending on the sample density at that location. Then, depending on the area covered by the sample, the variance of the Gaussian, $\sigma_m$, governing the beam model in Eq. (38), is calculated for each sample. As a result, the beam model automatically adapts to the local density of samples. Such a *location dependent* model results in a smooth likelihood function during global localization and a more peaked function during position tracking without changing the number of samples.

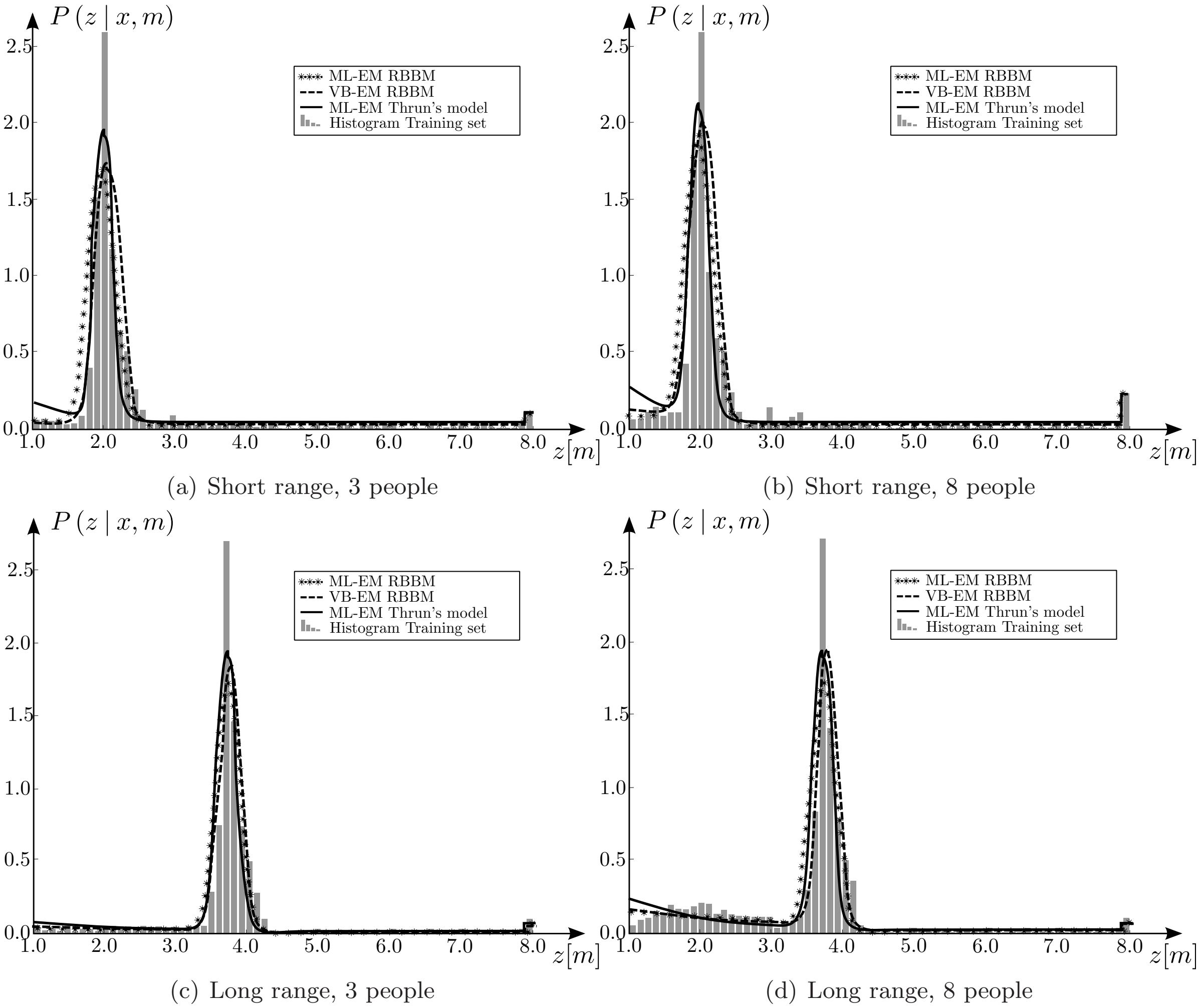

(a) Short range, 3 people

(b) Short range, 8 people

(c) Long range, 3 people

(d) Long range, 8 people

Figure 15: Comparison of the results of the ML-EM and VB-EM estimators for the RBBM and the results of a maximum likelihood estimator for two different expected ranges and two different number of people populating the robot environment.

Plagemann et al. (2007) and Pfaff et al. (2007) showed that by considering a region around samples, the individual beams become statistically dependent. The degree of dependency depends on the geometry of the environment and on the size and location of the considered region. Beam models, such as the RBBM, implicitly assume however that the beams are independent, that is:

$$P\left(\boldsymbol{z} \mid \boldsymbol{\theta}, x, m\right) \;=\; \prod_{b=1}^{B} P\left(z_b \mid \theta_b, x, m\right), \tag{104}$$

where $\boldsymbol{z} = \{z_b\}_{b=1:B}$ and $\boldsymbol{\theta} = \{\theta_b\}_{b=1:B}$ are the vectors containing the measured ranges and the angles of the different beams respectively; $z_b$ is the range measured at the beam with angle $\theta_b$; $B$ is the total number of beams and $P\left(z_b \mid \theta_b, x, m\right)$ is for instance the RBBM (Eq. (45)). By neglecting the dependency between beams, the resulting likelihoods $P\left(\boldsymbol{z} \mid \boldsymbol{\theta}, x, m\right)$ are overly peaked. Models taking into account the dependencies between beams consider the full range scan and are therefore called *full scan models* further on. The full scan models proposed by Plagemann et al. (2007) and Pfaff et al. (2007) both assume that the beams of a range scan are jointly Gaussian distributed. The off-diagonal elements of the covariance matrix associated with the Gaussian distribution represent the dependency. To learn the model parameters, both methods draw samples from the region around a sample and perform ray-casting using these samples. Plagemann et al. train a Gaussian process which models the full scan, while Pfaff et al. directly provide a maximum likelihood estimate for the mean and covariance of the Gaussian.

Section 6.1 shows that the dependency between beams may introduce multi-modality, even for simple static environments. Multi-variate Gaussian models as proposed by Plage-mann et al. (2007) and Pfaff et al. (2007) cannot handle this multi-modality. Therefore, a new sample-based method for obtaining an adaptive full scan model from a beam model, able to handle multi-modality, is proposed. Section 6.2 extends the adaptive full scan model for dynamic environments by taking into account non-Gaussian model uncertainty.

### 6.1 Sample-based Adaptive Full Scan Model for Static Environments

Plagemann et al. (2007) and Pfaff et al. (2007) estimate the full scan model, $P(\boldsymbol{z} \mid x, m)$[6], based on a local environment $\mathcal{U}(x)$ of the exact estimate $x$:

$$P(\boldsymbol{z} \mid x, m) \approx \int P(\tilde{x} \mid x)\, P_{\text{hit}}(\boldsymbol{z} \mid \tilde{x}, m)\, d\tilde{x}, \tag{105}$$

with $P(\tilde{x} \mid x)$ the distribution representing the probability that $\tilde{x}$ is an element of the environment $\mathcal{U}(x)$, i.e: $\tilde{x} \in \mathcal{U}(x)$. The environment $\mathcal{U}(x)$ is modeled as a circular region around x. Since this section does not consider the dynamics of the environment, only one component of the RBBM in Eq. (37) is used: $P_{\text{hit}}(z \mid x, m)$. The marginalization over the environment $\mathcal{U}(x)$ in Eq.(105) introduces dependencies between the measurements $z_b$ of the measurement vector $\boldsymbol{z}$.

The environment $\mathcal{U}(x)$, as explained above, depends on the sample density around the sample $x$ under consideration. Pfaff et al. (2006) proposed to use a circular region with diameter $d_{\mathcal{U}(x)}$, which is a weighted sum of the Euclidean distance and the angular difference. Like Plagemann et al. (2007) and Pfaff et al. (2007), an approximation of the above likelihood can be estimated online for each sample $x$ by simulating $L$ complete range scans at locations drawn from $\mathcal{U}(x)$ using the given map $m$ of the environment. Contrary to the multivariate Gaussian approximation proposed by Plagemann et al. and Pfaff et al., we propose a sample-based approximation, able to handle multi-modality. Sampling from the environment $\mathcal{U}(x)$ immediately results in a sample-based approximation of $P(\tilde{x} \mid x)$:

$$P(\tilde{x} \mid x) \approx \frac{1}{L} \sum_{l=1}^{L} \delta_{\tilde{x}^{(l)}}, \tag{106}$$

where $\delta_{\tilde{x}^{(l)}}$ denotes the delta-Dirac mass located in $\tilde{x}^{(l)}$, and the samples are distributed according to $P(\tilde{x} \mid x)$:

$$\tilde{x}^{(l)} \sim P(\tilde{x} \mid x). \tag{107}$$

Using this sample-based approximation of $P(\tilde{x} \mid x)$ the likelihood of Eq. (105) can be approximated as:

$$P(\boldsymbol{z} \mid x, m) \approx \frac{1}{L} \sum_{l=1}^{L} P_{\text{hit}}\left(\boldsymbol{z} \mid \tilde{x}^{(l)}, m\right). \tag{108}$$

Since this sample-based approximation has to be calculated online, the number of samples has to be limited. If the used environment $\mathcal{U}(x)$ is large, the resulting approximation will be

6. To simplify the notation $\boldsymbol{\theta}$ and $\theta$ are omitted from $P(\boldsymbol{z} \mid \boldsymbol{\theta}, x, m)$ and $P(z_b \mid \theta_b, x, m)$, respectively.

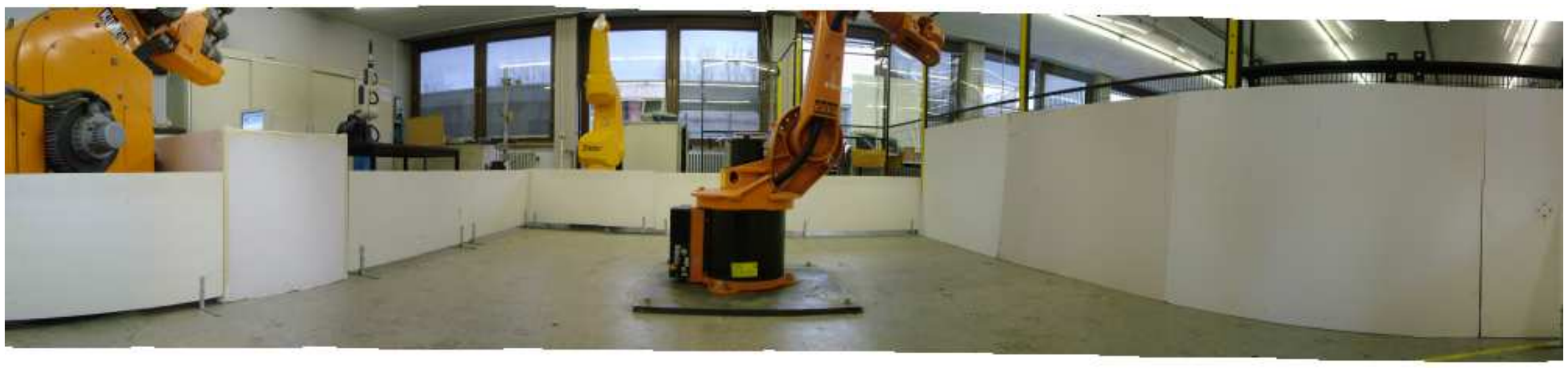

Figure 16: Panorama taken from the Sick LMS 200 range finder mounted on a Kuka 361 industrial robot. The environment consists of a rectangular 'room' with an object (a Kuka KR 15/2 robot) in the middle. We show that even for this simple static environment, the presented sample-based full scan model outperforms the Gaussian-based state of the art models.

bad. To smooth the undesired bumpy behavior due to the limited number of samples, the measurement noise $\sigma_m$ governing $P_{\text{hit}}(z \mid x, m)$ in Eq. (45), is artificially increased *depending on the size of* $\mathcal{U}(x)$ by multiplying it with a factor:

$$1 + C\sqrt{d_{\mathcal{U}(x)}}. \tag{109}$$

In further experiments, $C$ was set to 20.

#### 6.1.1 Experiment

A simple environment consisting of a rectangular 'room' with an object (a Kuka KR 15/2 robot) in the middle (Fig. 16) is used to show that the marginalization over (even small) $\mathcal{U}(x)$ to obtain the true likelihood not only introduces dependencies between the beams but also multi-modality. The $\mathcal{U}(x)$ results from a local uncertainty on the $x$- and $y-$position of $0.01m$ and a rotational uncertainty of $5°$. To obtain a reference, a Sick LMS 200 range finder is used to take a large number of measurements ($L = 1500$) at random locations sampled in $\mathcal{U}(x)$. To allow for exact positioning, the Sick LMS 200 is placed on a Kuka 361 industrial robot. The Sick LMS 200 range finder is connected to a laptop that controls the motion of the Kuka 361 industrial robot over the network using Corba-facilities in the Open Robot Control Software, Orocos (Bruyninckx, 2001; Soetens, 2006). A simplified map of the environment (Fig. 16) is built to simulate the 150 complete range scans needed to construct a full scan model. The marginal $P(z_b \mid x, m)$ of two selected beams are studied in more detail. The marginal likelihoods for the selected beam using the proposed sample-based approximation of Eq. (108) and the Gaussian approximation proposed by Pfaff et al. (2007), are compared in Fig. 17(b)-17(c). The histogram of the measurements of the selected beam in this figure clearly shows the multi-modality of the likelihood caused by the dependency between beams. In contrast to the Gaussian-based state of the art full scan model, the proposed sample-based approximation is able to handle the multi-modality of the range finder data. Fig. 17(d) shows the difference for all beams between the experimentally obtained cumulative marginal ($L = 1500$) and the Gaussian-based and sample-based approximation for all beams. The mean difference with the experimental data for the sample-based approximation is 2.8 times smaller than the difference for the Gaussian-based approximation, even for the simple static environment of Fig. 16 and the small $\mathcal{U}(x)$.

### 6.2 Sample-based Adaptive Full Scan Model for Dynamic Environments

The adaptive beam model proposed by Pfaff et al. (2006) is suited for use in dynamic environments since it uses the four component mixture beam model (Thrun et al., 2005; Choset et al., 2005). To date however, the adaptive full scan likelihood models of Pfaff et al. (2007) and Plagemann et al. (2007) have not been adapted for dynamic environments. The assumption that the beams are jointly Gaussian distributed, unable to capture the non-Gaussian uncertainty due to environment dynamics, prevents the straightforward extension for dynamic environments. In contrast, the sample-based approximation of the full scan likelihood, as proposed in Section 6.1, can be extended to include environment dynamics. To this end, replace $P_{\text{hit}}(z \mid x, m)$ in Eq. (105) and Eq. (108) by the full mixture of Eq. (38).

#### 6.2.1 Experiment

Fig. 18(a) and Fig. 18(b) compare the marginals for the selected beams (Fig. 17(a)) obtained from the adaptive full scan model for dynamic environments using the proposed sample-based approximation and the Gaussian approximation proposed by Pfaff et al. (2007). In contrast to the Gaussian-based state of the art full scan model, the proposed sample-based approximation is able to handle the multi-modality of the range finder data. Fig. 18(c) shows a probability map of the adaptive full scan model (sample-based approximation) suited for dynamic environments for the example environment of Fig. 16. The probability map plots $P(z \mid x, m)$ as a function of the position in the map and shows that the marginalization over the environment $\mathcal{U}(x)$ of a sample in Eq. (105) not only introduces dependency between beams but also introduces multi-modality.

## 7. Discussion

This paper proposed and experimentally validated the RBBM, a rigorously Bayesian network model of a range finder adapted to dynamic environments. All modeling assumptions are rigorously explained, and all model parameters have a physical interpretation. This approach resulted in a transparent and intuitive model. The rigorous modeling revealed all underlying assumptions and parameters. This way a clear physical interpretation of all parameters is obtained providing intuition for the parameter choices. In contrast to the model of Thrun et al. (2005), the assumption underlying the non-physical discontinuity in the RBBM is discovered. Furthermore, the paper proposes a different functional form for the probability of range measurements caused by unmodeled objects $P_{\text{occl}}(z \mid x, m)$ (Eq. (45)), i.e. quadratic rather than exponential as proposed by Thrun et al. Furthermore, compared to the work of Thrun et al. (2005), Choset et al. (2005), Pfaff et al. (2006) the RBBM depends on fewer parameters, while maintaining the same representational power for experimental data. Bayesian modeling revealed that both the rate of decay of $P_{\text{occl}}(z \mid x, m)$ and

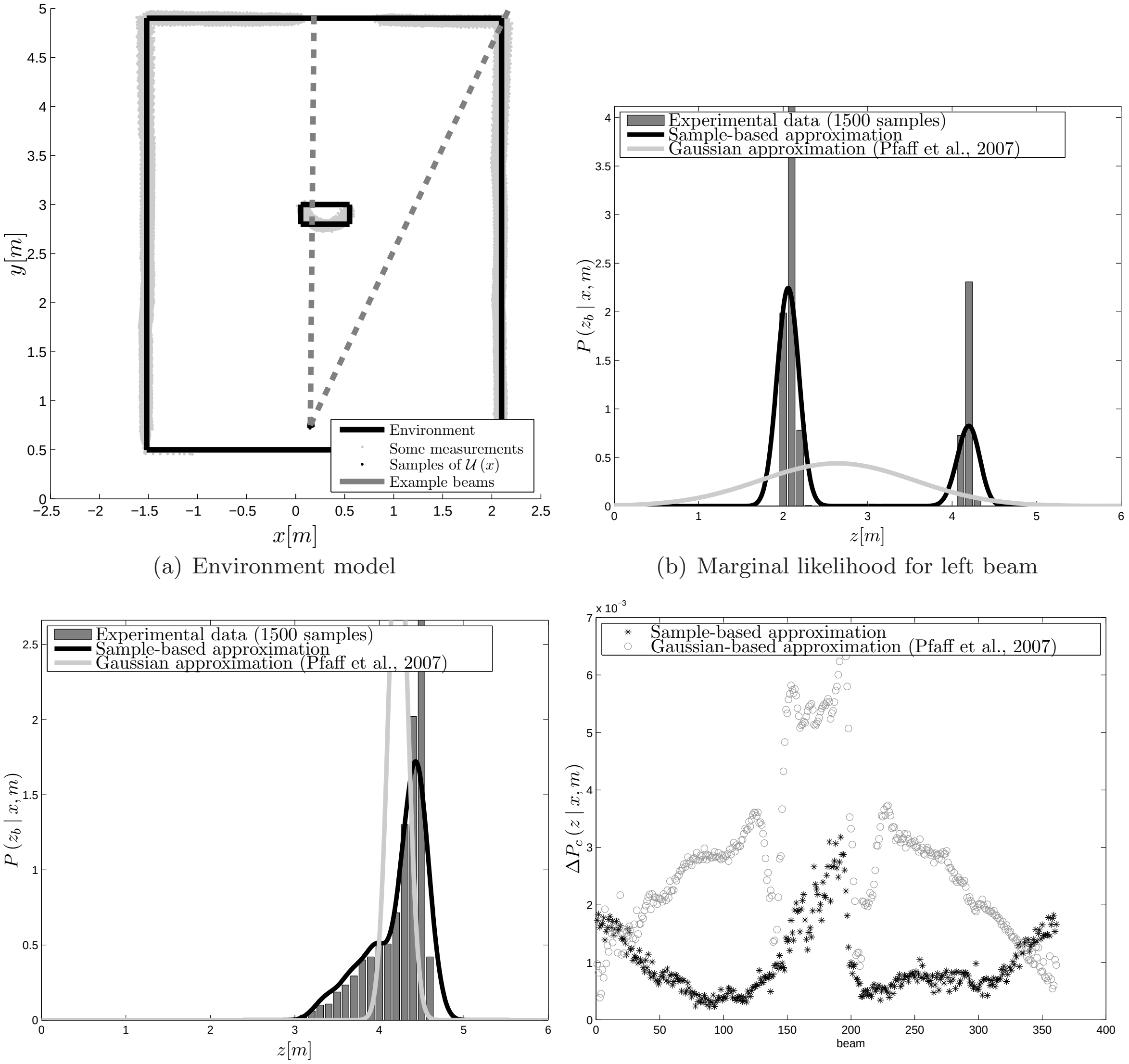

(a) Environment model

(b) Marginal likelihood for left beam

(c) Marginal likelihood for right beam

(d) Difference between experimental cumulative marginal and Gaussian and sample-based approximations

Figure 17: Experimental results for sample-based adaptive full scan model for static environments. (**a**) models the simple environment of Fig. 16. The range finder is located at $(0.15m, 0.75m)$. Samples from $\mathcal{U}(x)$ (resulting from a local uncertainty on the $x$- and $y-$position of $0.01m$ and a rotational uncertainty of $5°$) are shown with black dots, and some simulated measurements are shown in grey. (**b**) and (**c**) show the marginal likelihood $P(z_b \mid x, m)$ for the two selected beams together with the histogram of the experimentally recorded range finder data, the Gaussian-based approximation ($L = 150$) of Pfaff et al. (2007), and the sample-based approximation ($L = 150$) of this paper. (**d**) shows the difference for all beams between the experimentally obtained cumulative marginal ($L = 1500$) and the Gaussian-based and sample-based approximation.

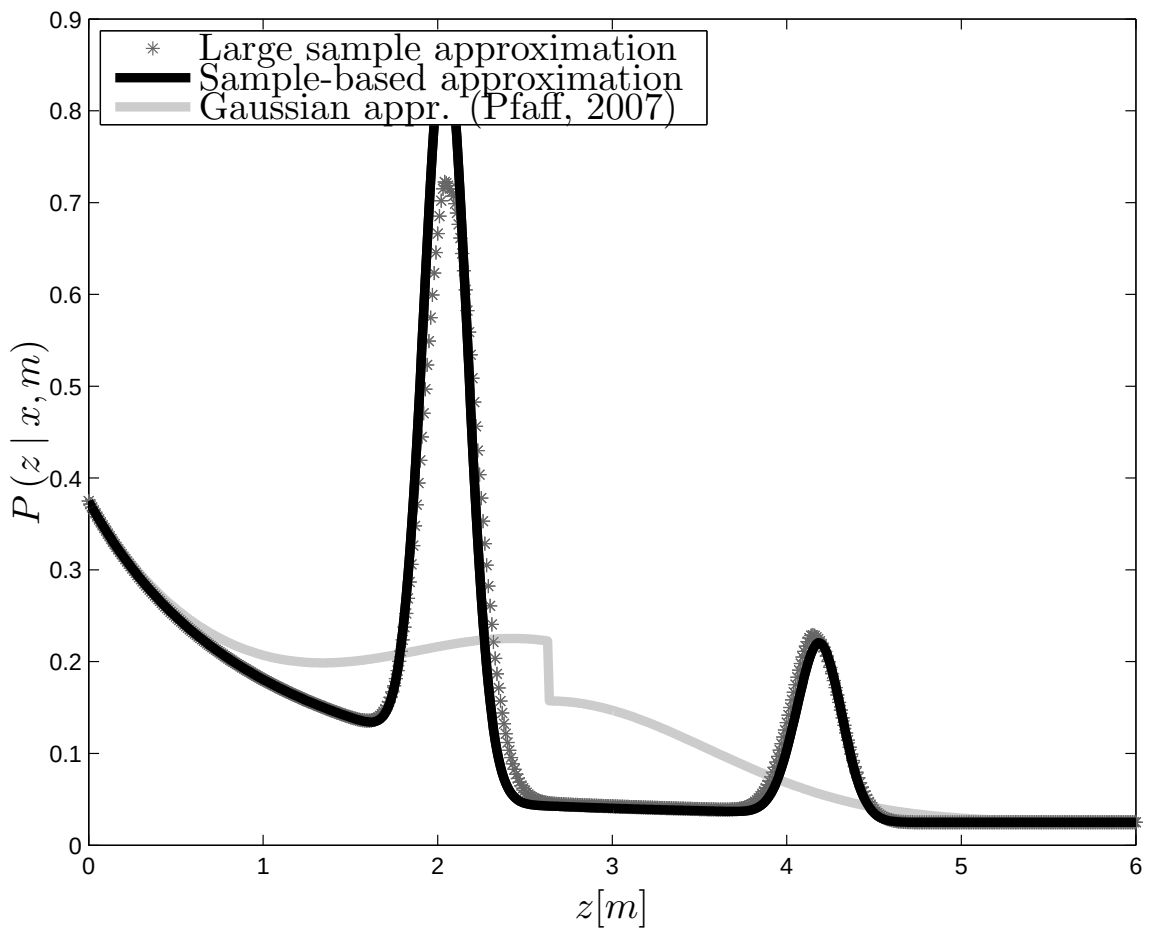

(a) Marginal likelihood for left beam

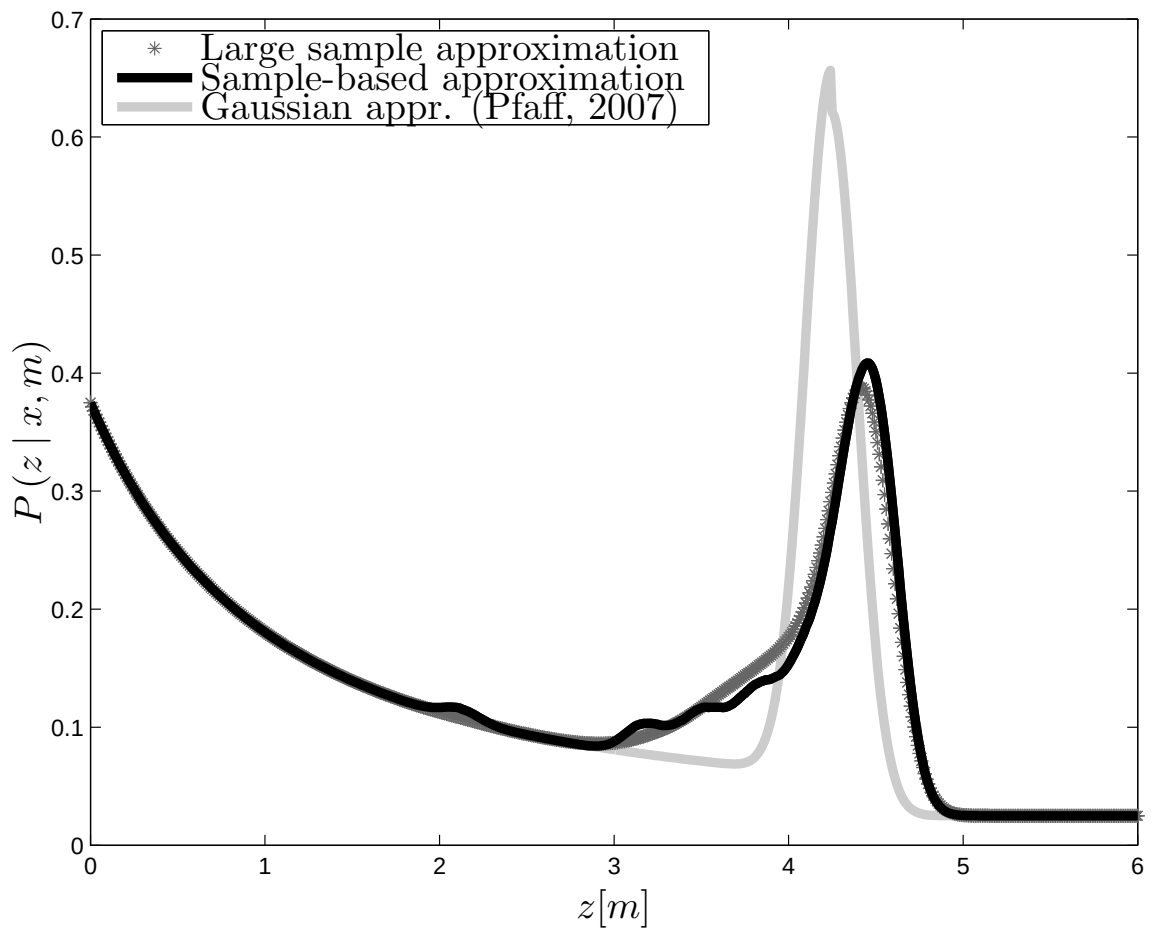

(b) Marginal likelihood for right beam

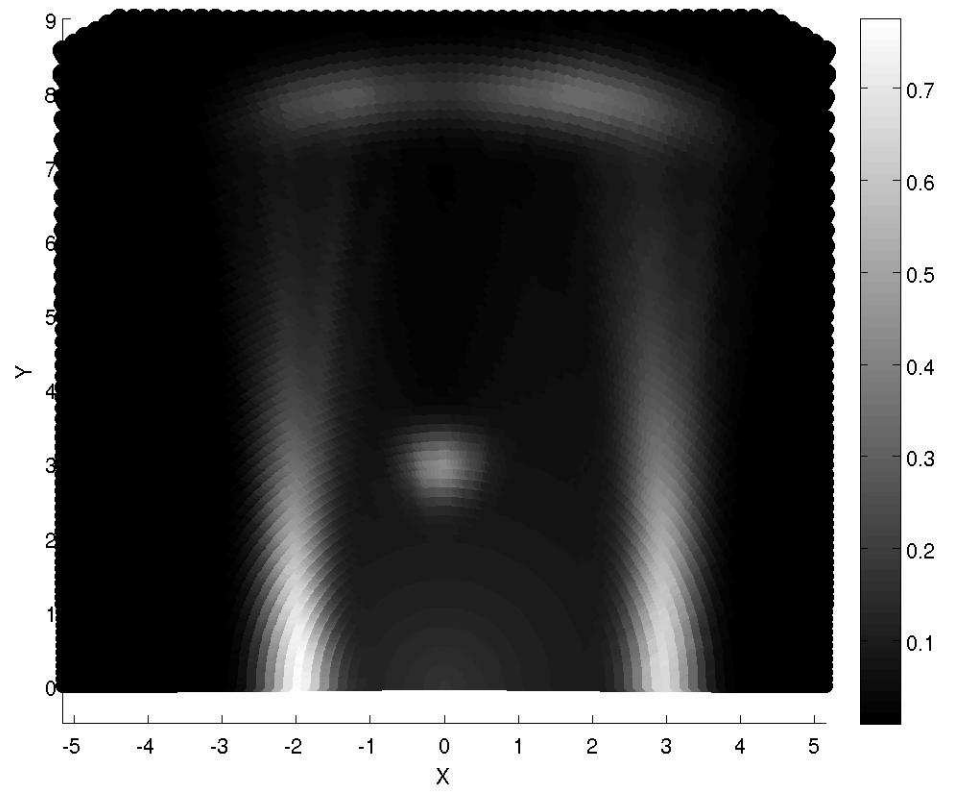

(c) Probability map $P(z \mid x, m)$ from sample-based approximation

Figure 18: Results for sample-based adaptive full scan model for dynamic environments. (**a**) and (**b**) show the marginal likelihood $P(z_b \mid x, m)$ for the two selected beams of Fig. 17(a) together with the Gaussian-based approximation ($L = 150$) of Pfaff et al. (2007) and the sample-based approximation ($L = 150$) extended for the use in dynamic environments. (**c**) shows the probability map resulting from the sample-based approximation. The probability map shows $P(z \mid x, m)$ as a function of the $x$- and $y-$ position in the map.

the probability of an occluded measurement $\pi_2$ depend on *one* parameter $p'$. State of the art sensor models however, assume independency of these two parameters. Finally, a maximum-likelihood and a variational Bayesian estimator (both based on expectation-maximization) were proposed to learn the model parameters of the RBBM. Learning the model parameters from experimental data benefits from the RBBM's reduced number of parameters. Using two sets of learning experiments from different application areas in robotics (one reported by Thrun et al. (2005)) the RBBM was shown to explain the obtained measurements at least as well as the state of the art model of Thrun et al.

Furthermore, the paper extended the RBBM to an adaptive full scan model in two steps: first, to a full scan model for static environments and next, to a full scan model for general, dynamic environments. The full scan model adapts to the local sample density when using a particle filter, and accounts for the dependency between beams. In contrast to the Gaussian-based state of the art models of Plagemann et al. (2007) and Pfaff et al. (2007), the proposed full scan model uses a sample-based approximation, which can cope with dynamic environments and with multi-modality (which was shown to occur even in simple static environments).

## Acknowledgments

The authors thank the anonymous reviewers for their thorough and constructive reviews. The authors also thank Wilm Decré, Pauwel Goethals, Goele Pipeleers, Ruben Smits, Bert Stallaert, Lieboud Van den Broeck, Marnix Volckaert and Hans Wambacq for participating in the experiments. All authors gratefully acknowledge the financial support by K.U.Leuven's Concerted Research Action GOA/05/10 and the Research Council K.U.Leuven, CoE EF/05/006 Optimization in Engineering (OPTEC). Tinne De Laet is a Doctoral Fellow of the Fund for Scientific Research–Flanders (F.W.O.) in Belgium.

## Appendix A. Simplification of Infinite Sum

To goal of this appendix is to prove that

$$\sum_n P(k \mid n, x, m) P(n) = \sum_{n=k}^{\infty} \left[ \begin{pmatrix} n \\ k \end{pmatrix} u^k (1-u)^{n-k} (1-p) p^n \right], \tag{110}$$

can be simplified to:

$$\sum_n P(k \mid n, x, m) P(n) = \left(1 - p'\right) p'^k, \tag{111}$$

with $p' = \frac{up}{1-(1-u)p}$.

Expand Eq. (110) and move terms out of the summation so that:

$$\sum_n P(k \mid n, x, m) P(n) = (1-p) u^k p^k \sum_{n=k}^{\infty} \left[ \frac{n!}{(n-k)!k!} \left[(1-u) p\right]^{n-k} \right]. \tag{112}$$

Next introduce variables $t = n - k$ and a $e = (1-u) p$:

$$\sum_n P(k \mid n, x, m) P(n) = (1-p) u^k p^k \sum_{t=0}^{\infty} \left[ \frac{(t+k)!}{t!k!} e^t \right]. \tag{113}$$

The next step is to prove by induction that:

$$\sum_{t=0}^{\infty}\left[\frac{(t+k)!}{t!k!}e^{t}\right]=\frac{1}{(1-e)^{k+1}}. \tag{114}$$

First, show that the above equality holds for $k=0$:

$$\sum_{t=0}^{\infty}\frac{t!}{t!}e^{t}=\sum_{t=0}^{\infty}e^{t}, \tag{115}$$

which is the well-known geometric series, so:

$$\sum_{t=0}^{\infty}e^{t}=\frac{1}{1-e}, \tag{116}$$

showing that equality (114) indeed holds for $k=0$. Next it is proved that, if the expression holds for $k-1$, it also holds for $k$. Introduce variable $V$ for the solution of the infinity sum for $k$ and split up (114) in two parts:

$$V=\sum_{t=0}^{\infty}\left[\frac{(t+k)!}{t!k!}e^{t}\right]=\sum_{t=0}^{\infty}\left[\frac{(t+k)!}{t!k!}-\frac{(t+k-1)!}{t!\,(k-1)!}\right]e^{t}+\underbrace{\sum_{t=0}^{\infty}\left[\frac{(t+k-1)!}{t!\,(k-1)!}e^{t}\right]}_{\frac{1}{(1-e)^{k}}}, \tag{117}$$

where the fact is used that the equality (114) holds for $k-1$. Simplify the first term of the summation to:

$$\sum_{t=0}^{\infty}\left[\frac{(t+k)!}{t!k!}-\frac{(t+k-1)!}{t!\,(k-1)!}\right]e^{t}=\sum_{t=0}^{\infty}\left[\frac{(t+k-1)!}{t!k!}te^{t}\right]. \tag{118}$$

The term in the summation for $t=0$ is equal to zero. Hence, introduce the variable $t'=t-1$ and simplify:

$$\begin{aligned}\sum_{t=0}^{\infty}\left[\frac{(t+k)!}{t!k!}-\frac{(t+k-1)!}{t!\,(k-1)!}\right]e^{t} &= \sum_{t'=0}^{\infty}\left[\frac{(t'+k)!}{(t'+1)!k!}\left(t'+1\right)e^{t'+1}\right] && (119)\\ &= e\underbrace{\sum_{t'=0}^{\infty}\left[\frac{(t'+k)!}{t'!k!}e^{t'}\right]}_{V}, && (120)\end{aligned}$$

from which the series $V$ we were looking for is recognized. Substitute the above result in Eq. (117) so that:

$$V=eV+\frac{1}{(1-e)^{k}}. \tag{121}$$

Solving this equation for V gives:

$$V=\frac{1}{(1-e)^{k+1}}, \tag{122}$$

proving that equality (114) holds for $k$ if it assumed to hold for $k-1$, and closing the proof by induction.

Now substitute Eq. (114) in Eq. (113):

$$\sum_n P\left(k \mid n, x, m\right) P\left(n\right) = \frac{\left(1-p\right) u^k p^k}{\left[1-\left(1-u\right)p\right]^{k+1}}, \tag{123}$$

or rewrite as:

$$\sum_n P\left(k \mid n, x, m\right) P\left(n\right) = \left(1-p'\right) p'^k, \tag{124}$$

where $p' = \frac{up}{1-(1-u)p}$, what is exactly what had to be proved.